\documentclass[10pt,twocolumn,letterpaper]{article}

\usepackage{cvpr}
\usepackage{times}
\usepackage{epsfig}
\usepackage{graphicx}
\usepackage{amsmath}
\usepackage{amssymb}
\usepackage{algorithm} 
\usepackage{algpseudocode}

\algrenewcommand\algorithmicrequire{\textbf{Input:}}
\algrenewcommand\algorithmicensure{\textbf{Output:}}

\algnewcommand{\Initialize}[1]{%
	\State \textbf{Initialize:}
	\Statex \hspace*{\algorithmicindent}\parbox[t]{.8\linewidth}{\raggedright #1}
}

\renewcommand{\mathbf}{\boldsymbol}
\newcommand{\LL}{\mathbf{L}}
\renewcommand{\P}{\mathbf{P}}
\newcommand{\M}{\mathcal{M}}

\renewcommand{\l}{\mathbf{l}}

\newcommand{\rr}{\mathbf{r}}

\algnewcommand\And{\textbf{and}}
\algnewcommand\Or{\textbf{or}}

\algnewcommand{\IIf}[1]{\State\algorithmicif\ #1\ \algorithmicthen}
\algnewcommand{\EndIIf}{\unskip\ \algorithmicend\ \algorithmicif}

\usepackage[pagebackref=true,breaklinks=true,letterpaper=true,colorlinks,bookmarks=false]{hyperref}

\def\cvprPaperID{1446 \vspace{-2mm}} 

\cvprfinalcopy 

\renewcommand{\mathbf}{\boldsymbol}
\renewcommand{\L}{\mathbf{L}}

\newcommand{\W}{\mathbf{W}}

\newcommand{\V}{\mathcal{V}}

\renewcommand{\P}{\mathbf{P}}

\newcommand{\p}{\mathbf{p}}
\newcommand{\q}{\mathbf{q}}
\renewcommand{\l}{\mathbf{l}}

\newcommand{\x}{\mathbf{x}}

\newcommand{\nop}[1]{}

\begin{document}

\title{Learning to Parse Wireframes in Images of Man-Made Environments \vspace{-4mm}}

\author{Kun Huang$^1$, Yifan Wang$^1$, Zihan Zhou$^2$, Tianjiao Ding$^1$, Shenghua Gao$^1$, and Yi Ma$^3$\\
$^1$ShanghaiTech University {\tt\small \{huangkun, wangyf, dingtj, gaoshh\}@shanghaitech.edu.cn}\\
$^2$The Pennsylvania State University {\tt\small zzhou@ist.psu.edu}\\
$^3$University of California, Berkeley {\tt\small yima@eecs.berkeley.edu}
}

\maketitle

\begin{abstract}
In this paper, we propose a learning-based approach to the task of automatically extracting a ``wireframe'' representation
for images of cluttered man-made environments. The wireframe (see Fig.~\ref{fig:teaser}) contains all salient straight lines
and their junctions of the scene that encode efficiently and accurately large-scale geometry and object shapes. To this end, we have built
a very large new dataset of over 5,000 images with wireframes thoroughly labelled by humans. We have proposed two convolutional neural networks
that are suitable for extracting junctions and lines with large spatial support, respectively. The networks trained on our dataset
have achieved significantly better performance than state-of-the-art methods for junction detection and line segment
detection, respectively. We have conducted extensive experiments to evaluate quantitatively and qualitatively the wireframes
obtained by our method, and have convincingly shown that effectively and efficiently parsing wireframes for images of
man-made environments is a feasible goal within reach. Such wireframes could benefit many important visual tasks such as
feature correspondence, 3D reconstruction, vision-based mapping, localization, and navigation. The data and source code are available at \url{https://github.com/huangkuns/wireframe}.
\end{abstract}


\section{Introduction}
How to infer 3D geometric information of a scene from 2D images has been a fundamental problem in computer vision. Conventional approaches to build a 3D model typically rely on detecting, matching, and triangulating local image features (e.g. corners, edges, SIFT features, and patches). One great advantage of working with local features is that the system can be somewhat oblivious to the scene, as long as it contains sufficient distinguishable features. Meanwhile, modern applications of computer vision systems often require an autonomous agent (e.g., a car, a robot, or a UAV) to efficiently and effectively negotiate with a physical space in cluttered {\em man-made} (indoor or outdoor) environments. Such scenarios present significant challenges to the current local-feature based approaches: Man-made environments typically consist of large textureless surfaces (e.g. white walls or the ground); or they may be full of repetitive patterns hence local features are ambiguous to match; and the visual localization system is required to work robustly and accurately over extended routes and sometimes across very large baseline between views. 

\nop{
\begin{figure}[t]
\centering
\includegraphics[width= 0.4\textwidth]{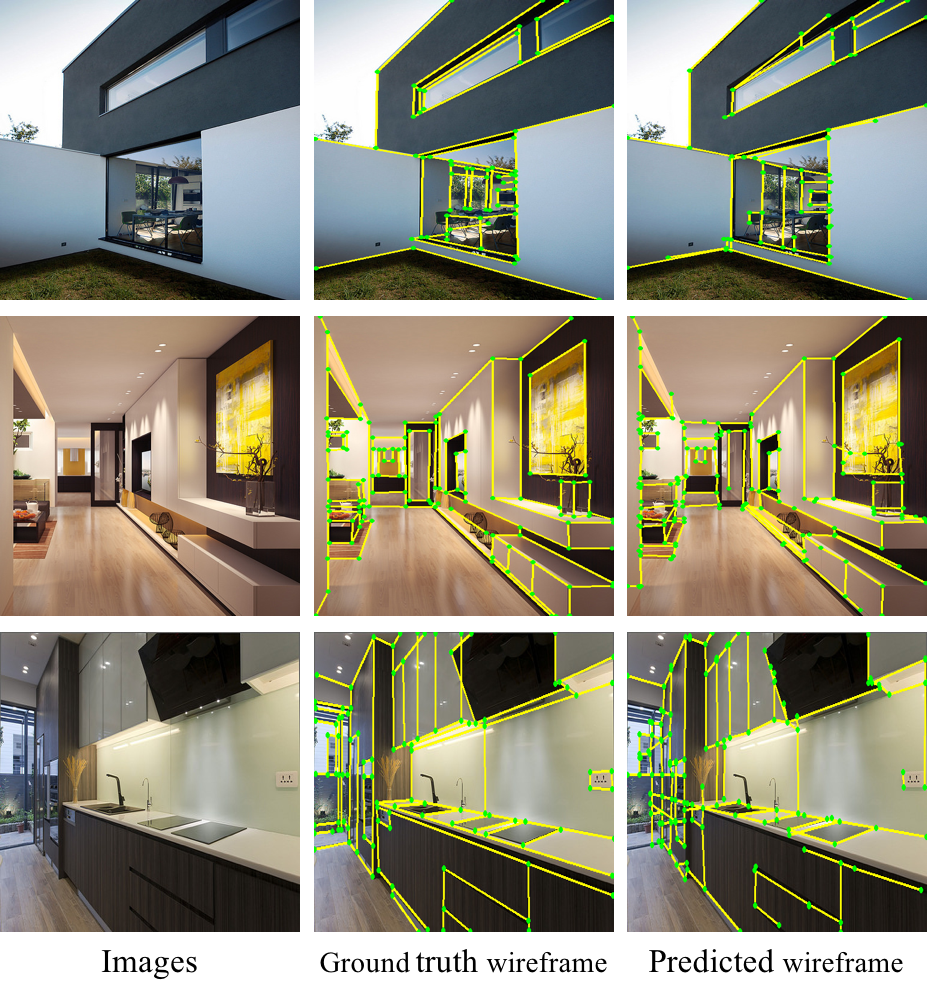}\vspace{-2mm}
\caption{ \vspace{-5mm}}
\label{fig:teaser}
\end{figure}
}

\begin{figure}
\centering
\includegraphics[height= 1.03in]{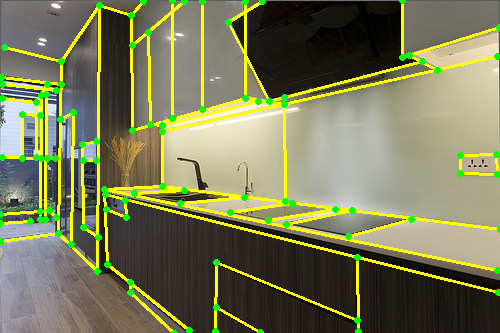}
\includegraphics[height= 1.03in]{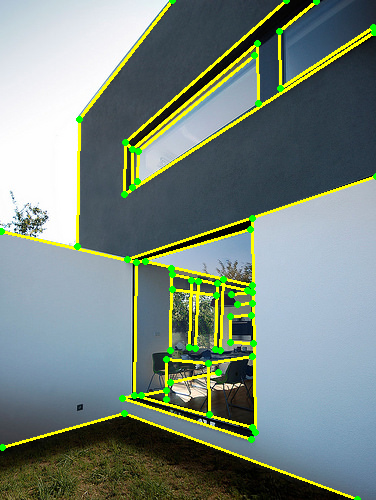}
\includegraphics[height= 1.03in]{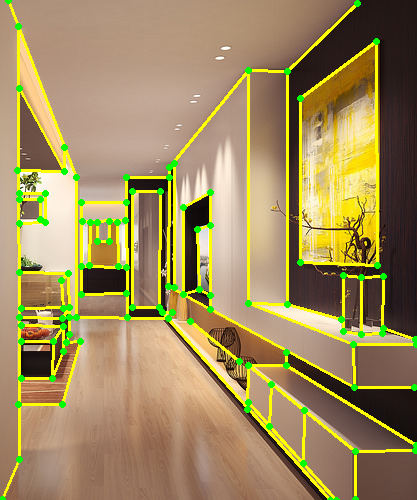}\\
\vspace{0.5mm}\includegraphics[height= 1.03in]{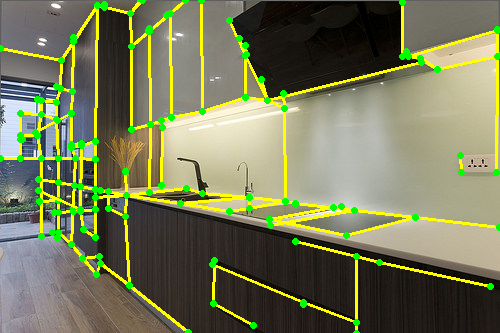}
\includegraphics[height= 1.03in]{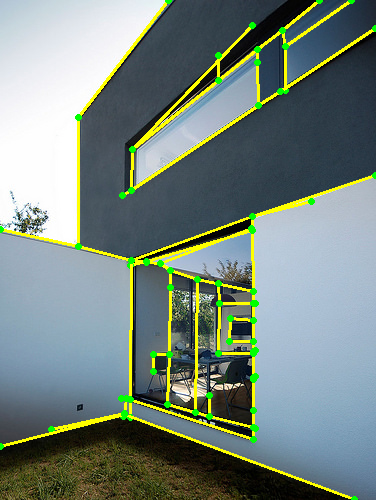}
\includegraphics[height= 1.03in]{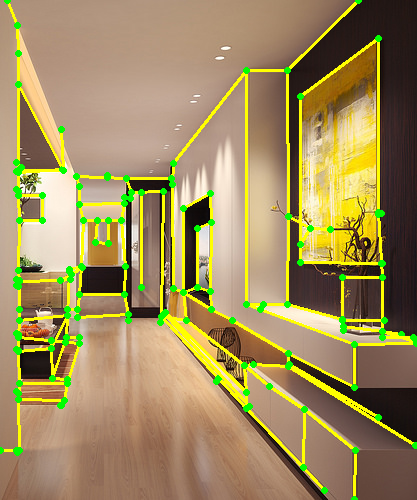}\\
\caption{{\bf First row:} Examples of typical indoor or outdoor scenes with geometrically meaningful wireframes labelled by humans; {\bf Second row:} Wireframes automatically extracted by our method.}
\vspace{-3mm}
\label{fig:teaser}
\end{figure}

Nevertheless, the human vision system seems capable of effortlessly localizing or navigating among such environments arguably by exploiting larger-scale (global or semi-global) structural features or regularities of the scene. For instance, many works \cite{DelageLN05, LeeHK09, FurukawaCSS09, FlintMR11, RamalingamPJT13} have demonstrated that prior knowledge about the scene such as a Manhattan world could significantly benefit the 3D reconstruction tasks. The Manhattan assumption can often be violated in cluttered man-made environments, but it is rather safe to assume that man-made environments are dominantly piece-wise planar hence rich of visually salient lines (intersection of planes) and junctions (intersection of lines). Conceptually, such junctions or lines could just be a very small ``subset'' among the local corner features (or SIFTs) and edge features detected by conventional methods, but they already encode most information about larger-scale geometry of the scene. For simplicity, we refer to such a set of lines and their intersected junctions collectively as a ``wireframe''.\footnote{In architecture design, a wireframe is often referred to a line drawing of a building or a scene on paper. Interpretation of such line drawings of 3D objects has a long history in computer vision dated back to the '70s and '80s~\cite{Huffman71, Clowes71, Sugihara82, Malik87}.}

The goal of this work is to {\em study the feasibility of developing a vision system that could efficiently and effectively extract the wireframe of a man-made scene}. Intuitively, we wish such a system could emulate the level of human perception of the scene geometry, even from a single image. To this end, we have collected a database of over 5,000 images of typical indoor and outdoor environments and asked human subjects to manually label out all line segments and junctions that they believe to be important for understanding shape of regular objects or global geometric layout of the scene.\footnote{For simplicity, this work is limited to wireframes consisting of straight lines. But the idea and method obviously apply to wireframes with curves.} Fig. \ref{fig:teaser} (first row) shows some representative examples of labelled wireframes.

In the literature, several methods have been proposed to detect line segments~\cite{von2012lsd} or junctions~\cite{RamalingamPJT13,XiaDG14} in the image, and to further reason about the 3D scene geometry using the detected features~\cite{LeeHK09, FlintMR11, ElqurshE11, RamalingamB13, Yang16}. These methods typically take a bottom-up approach: First, line segments are detected in the image. Then, two or more segments are grouped to form candidate junctions. However, there are several inherent difficulties with this approach. \emph{First}, by enumerating all pairs of line segments, a large number of intersections are created. But only a very small subset of them are true junctions in the image. To retrieve the true junctions, various heuristics as well as RANSAC-based verification techniques have been previously proposed. As result, such methods are often time consuming and even break down when the scene geometry and texture become complex. \emph{Second}, detecting line segments itself is a difficult problem in computer vision. If one fails to detect certain line segments in the image, then it would be impossible for the method to find the associated junctions. \emph{Third}, since all existing methods rely on low-level cues such as image gradients and edge features to detect line segments and junctions, they are generally unable to distinguish junctions and line segments that are of global geometric importance with those produced by local textures or irregular shapes.


In view of the fundamental difficulties of existing methods, we propose a complementary approach to wireframe (junctions and line segments) detection in this paper. Our method does not rely on grouping low-level features such as image gradients and edges. Instead, we directly learn detectors for junctions and lines of large spatial support from the above large-scale dataset of manually labeled junctions and lines. In particular, inspired by the recent success of convolutional neural networks in object detection, we design novel network architectures for junction and line detection, respectively. We then give a simple but effective method to establish incidence relationships among the detected junctions and lines and produce a complete wireframe for the scene. Fig. \ref{fig:teaser} (second row) shows typical results of the proposed method. As one can see, our method is able to detect junctions formed by long line segments with weak gradient while significantly reducing the number of false detections. In addition, as the labelled junctions and line segments are primarily associated with salient, large-scale geometric structures of the scene, the resulting wireframe is geometrically more meaningful, emulating human perception of the scene's geometry.

\noindent{\bf Contributions of this work} include: (i) the establishment of a large dataset for learning-based wireframe detection of man-made environments, and (ii) the development of effective, end-to-end trainable CNNs for detecting geometrically informative junctions and line segments. Comparing with existing methods on junction and line segment detection, our learning-based method has achieved, both quantitatively and qualitatively, superior performances on both tasks, hence convincingly verified the feasibility of wireframe parsing. Furthermore, both junction and line detection achieves almost real-time performance at the testing time, thus is suitable for a wide range of real-world applications such as feature correspondence, 3D reconstruction, vision-based mapping, localization and navigation.

\subsection*{Related Work}
\vspace{-2mm}
\noindent{\bf Edge and line segment detection.} Much work has been done to extract line segments from images. Existing methods are typically based on perceptual grouping of low-level cues (i.e., image gradients)~\cite{NietoCSG11, von2012lsd, BrownWG15, LiuCGNZT15, LuYLL15}. A key challenge of these local approaches is the choice of some appropriate threshold to discriminate true line segments from false conjunctions. Another line of work extends Hough transform to line segment detection~\cite{MatasGK00, FurukawaS03, XuSK15}. While Hough transform has the ability to accumulate information over the entire image to determine the presence of a line structure, identifying endpoints of the line segment in the image remains a challenge~\cite{Almazan17}. Recently, machine learning based approaches have been shown to produce the state-of-the-art results in generating pixel-wise edge maps~\cite{DollarZ13, XieT15, ManinisPAG16}. But these methods do not attempt to extract straight line segments from the image.

\noindent{\bf Junction detection.} 
Detecting and analyzing junctions in real-world images remains a challenging problem due to a large number of fragmented, spurious, and missing line segments. In the literature, there are typically two ways to tackle this problem. The first group of methods focuses on operators based on local image cues, such as the Harris corner detector~\cite{HarrisS88}. However, local junction detection is known to be difficult, even for humans~\cite{McDermott04}. More recent methods detect junctions by first locating contours (in natural images)~\cite{MaireAFM08} or straight line segments (in man-made environments)~\cite{LeeHK09, RamalingamPJT13, ElqurshE11, XiaDG14} and then grouping them to form junctions. As we discussed before, such bottom-up methods are (i) sensitive to scene complexity, and (ii) vulnerable to imperfect line segment detection results.

\noindent{\bf Line- and junction-based geometric reasoning.} Knowledge about junctions and the associated line structures is known to benefit many real-world 3D vision tasks. From a single image, a series of recent work use these features to recover the 3D layout of the scene~\cite{LeeHK09, RamalingamPJT13, RamalingamB13, Yang16}. Meanwhile, observing that the junctions impose incident constraints on the adjacent line segments, \cite{JainKTS10} devises a method for 3D reconstruction of lines without explicitly matching them across views, whereas \cite{WangFSZLCTQ16} proposes a surface scaffold structure that consists of sets of connected edges to regularize stereo-based methods for building reconstruction. Furthermore, \cite{ElqurshE11} uses line segments and junctions to develop a robust and efficient method for two-view pose estimation, and \cite{Xu2017} systematically studies how knowledge about junctions can affect the complexity and number of solutions to the Perspective-n-Line (PnL) problem.

\noindent{\bf Machine learning and geometry.} There is a large body of work on machine learning based approach to inferring pixel-level geometric properties of the scene, such as the depth~\cite{SaxenaCN08, EigenPF14}, and the surface normal~\cite{FouheyGH13, FouheyGH14}. But few work has been done on detecting mid/high-level geometric primitives with supervised training data. Recently, \cite{HainesC15} proposes a method to recognize planes in a single image, \cite{GuoZH15} uses SVM to classify indoor planes (e.g., walls and floors), and \cite{MallyaL15, RenCLK16a, DasguptaFCS16} train fully convolutional networks (FCNs) to predict ``informative'' edges formed by the pairwise intersections of room faces. However, none of the work aims to detect highly compressive vectorized junctions or line segments in the image, let alone a complete wireframe.

\section{A New Dataset for Wireframe Detection}

\begin{figure}[t]
\centering
\includegraphics[height= 0.95in]{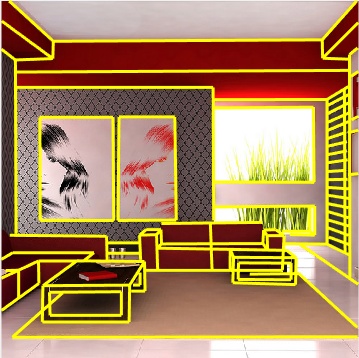}
\includegraphics[height= 0.95in]{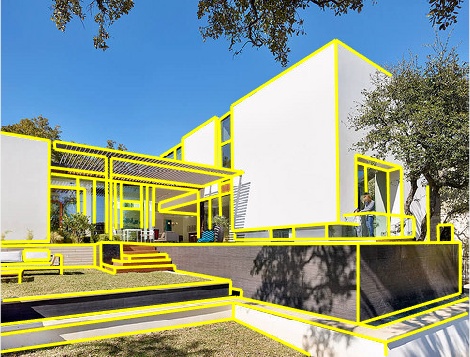}
\includegraphics[height= 0.95in]{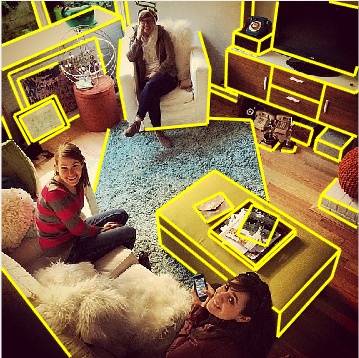}\\
\vspace{0.5mm}\includegraphics[height= 0.95in]{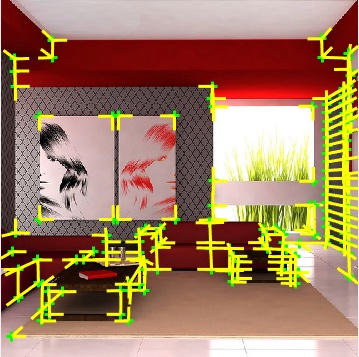}
\includegraphics[height= 0.95in]{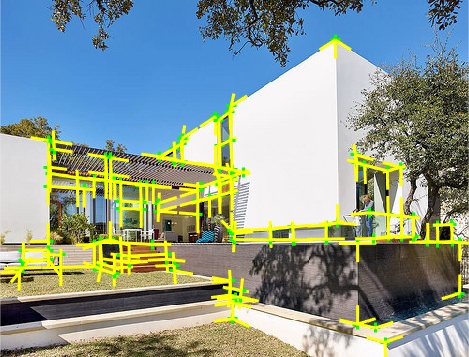}
\includegraphics[height= 0.95in]{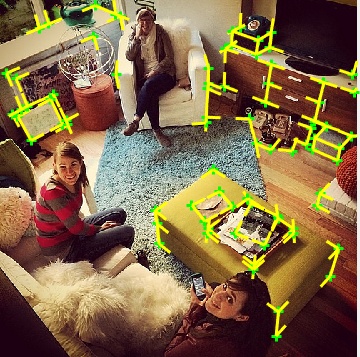}
\caption{Example images of our wireframe dataset, which covers a wide range of man-made scenes with different viewpoints, lighting conditions, and styles. For each image, we show the manually labelled line segments (first row) and the ground truth junctions derived from the line segments (second row).\vspace{-4mm}}
\label{fig:data}
\end{figure}


As part of our learning-based framework to wireframe detection, we have collected 5,462 images of man-made environments. Some examples are shown in Fig.~\ref{fig:data}. The scenes include both indoor environments such as bedroom, living room, and kitchen, and outdoor scenes, such as house and yard. For each image, we manually labelled all the line segments associated with the scene structures. Here, our focus is on the \emph{structural elements} in the image, that is, elements (i.e., line segments) from which meaningful geometric information of the scene can be extracted. As a result, we do not label line segments that are associated with texture (e.g., curtains, tree leaves), irregular or curved objects (e.g., sofa, humans, plants), shadows etc.

With the labelled line segments, ground truth junction locations and their branches can be easily obtained from the intersection or incidence relationships among two or more line segments in the image (Fig.~\ref{fig:data}, second row). Note that, unlike previous works~\cite{RamalingamPJT13, RamalingamB13}, we do not restrict ourselves to \emph{Manhattan junctions}, which are formed by line segments aligned with one of three principal and mutually orthogonal directions in the scene. In fact, many scenes in our dataset do not satisfy the Manhattan world assumption~\cite{CoughlanY03}. For example, the scene depicted in the last column of Fig.~\ref{fig:data} has more than two horizontal directions.

In summary, our annotation in each image includes a set of {\em junction points} $\mathbf{P} = \{\p_n\}_{n=1}^N$ and a set of {\em line segments} $\L = \{\l_m\}_{m=1}^M$. Each junction $\p$ is the intersection of several, say $R$, line segments, called its branches. The coordinates of $\p$ are denoted as $\x \in \mathbb{R}^2$ and its line branches are recorded by their angles $\{\theta_{r}\}_{r=1}^R$. The number $R$ is known as the order of the junction, and the typical ``$L$'', ``$Y$'', and ``$X$''-type junctions have orders $R= 2$, $3$, and $4$, respectively. Each line segment is represented by its two end points: $\l = (\p_{1}, \p_{2})$. Hence, the {\em wireframe}, denoted as $\W$, records all incidence and intersection relationships between junctions in $\P$ and lines in $\L$. It can be represented by an $N \times M$ matrix $\mathbf{W}$ whose $nm$-th entry is 1 if $\p_n$ is on $\l_m$, and 0 otherwise. Notice that two line segments are intersected at some junction if and only if the corresponding entry in $\W^T \W$  is nonzero; and similarly $\W\W^T$ for connected junctions.

\section{Wireframe Detection Method}

\begin{figure*}[t]
	\centering
	\includegraphics[width=0.95\textwidth]{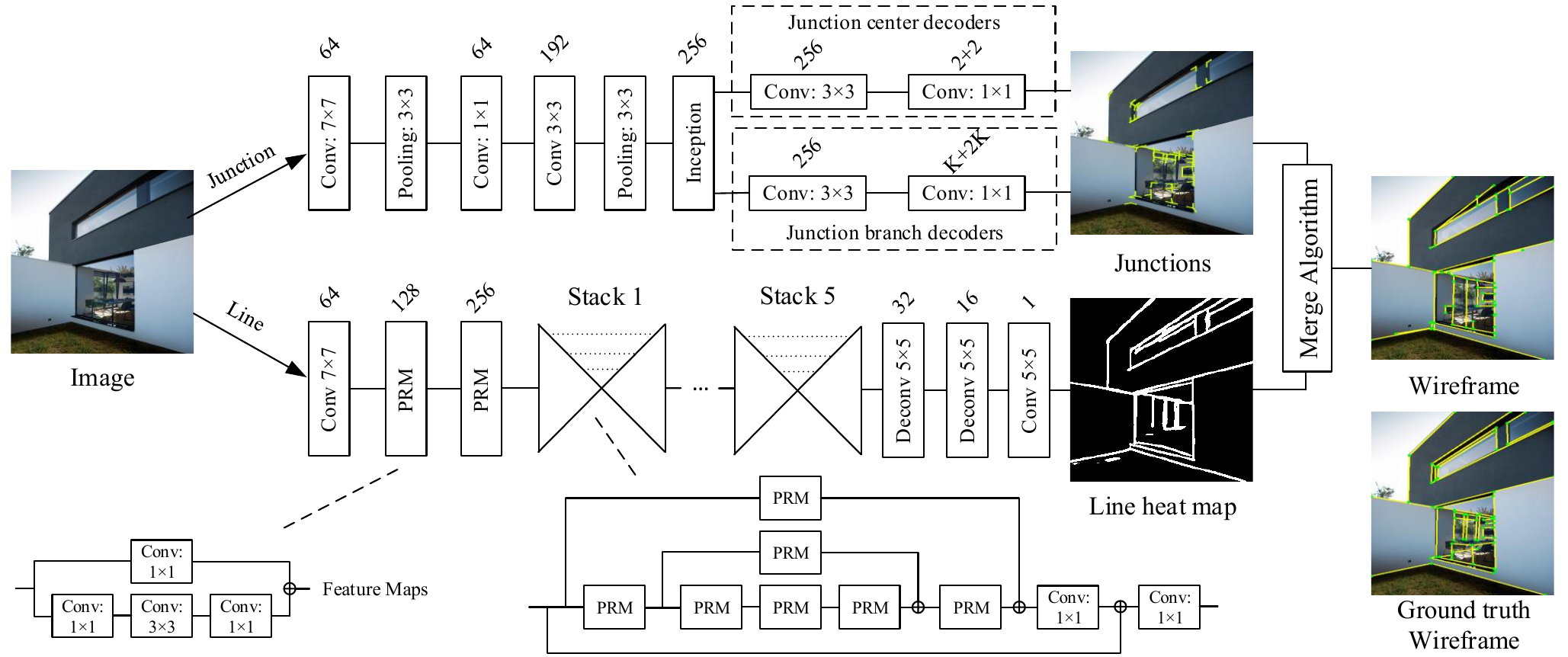}
	\caption{Architecture of the overall system. {\bf Top:} junction detection network. {\bf Bottom:} line detection network.} \vspace{-5mm}
	\label{fig:system}
\end{figure*}

Recently, deep convolutional neural networks (CNNs) such as~\cite{SermanetEZMFL13, RenHGS15, RedmonF16} have shown impressive performance in object detection tasks. Utilizing the dataset we have, we here design new, end-to-end trainable CNNs for detecting junctions and lines, respectively, and then merge them into a complete wireframe. Fig.~\ref{fig:system} shows the overall architecture of our proposed networks and method. Note that we choose different network architectures for junctions and lines due to the nature of their geometric properties, which we will elaborate below.

\subsection{Junction Detection}

\subsubsection{Design Rationale} \vspace{-2mm}

\begin{figure}[t]
\centering
\includegraphics[width=0.35\textwidth]{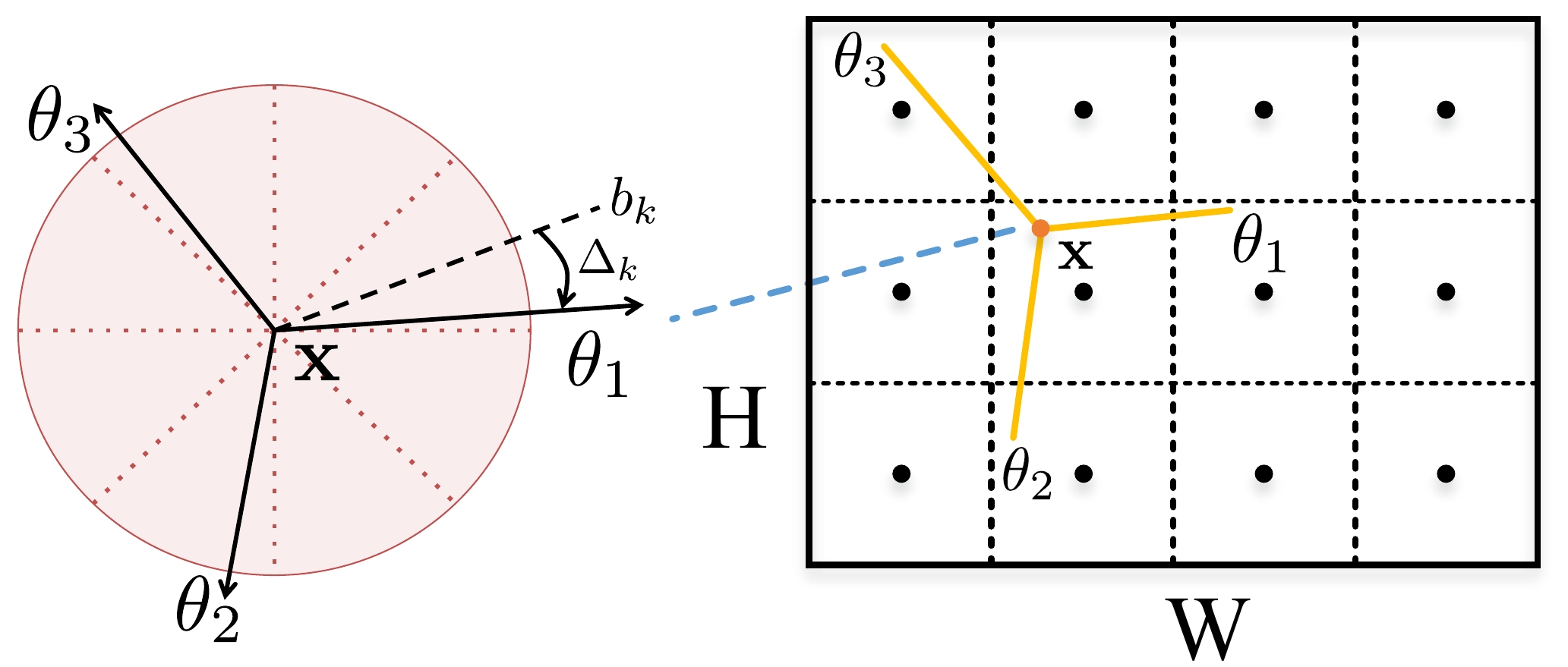}\vspace{-1mm}
\caption{Representation of a junction with three branches. \vspace{-5mm}}
\label{fig:model}
\end{figure}

Our design of the network architecture is guided by several important observations about junctions.

\noindent{\bf Fully convolutional network for global detection.} As we mentioned before, \emph{local} junction detection is a difficult task, which often leads to spurious detections. Therefore, it is important to enable the network to reason globally when making predictions. This motivates us to choose a \emph{fully convolutional network} (FCN), following its recent success in object detection~\cite{RedmonF16}. Unlike other popular object detection techniques that are based on sliding windows~\cite{SermanetEZMFL13} or region proposals~\cite{RenHGS15}, FCN sees the entire image so it implicitly captures the contextual information about the junctions. 
Similar to~\cite{RedmonDGF16, RedmonF16}, our network divides the input image into an $H\times W$ mesh grid, see Fig.~\ref{fig:model} right. If the center of a junction falls into a grid cell, that cell is responsible for detecting it. Thus, each $ij$-th cell predicts a confidence score $c_{ij}$ reflecting how confident the model thinks there exists a junction in that cell. To further locate the junction, each $ij$-th cell also predicts its relative displacement $\x_{ij}$ w.r.t. the center of the cell. Note that the behavior of the grid cells resembles the so-called ``anchors'', which serve as regression references in the latest object detection pipelines~\cite{RenHGS15, RedmonDGF16, Liu2016SSD}.

\noindent{\bf Multi-bin representation for junction branches.} Unlike traditional object detection tasks, each cell in our network needs to make different numbers of predictions due to the varying number of branches in a junction. To address this issue, we borrow the idea of spatial grid and propose a new multi-bin representation for the branches, as shown in Fig.~\ref{fig:model} left. We divide the circle (i.e., from 0 to 360 degrees) into $K$ equal bins, with each bin spanning $ \frac{360}{K}$ degrees. Let the center of the $k$-th bin be $b_k$, we then represent an angle $\theta$ as $(k, \Delta_k)$, if $\theta$ fall into the $k$-th bin, where $\Delta_k$ is the angle residual from the center $b_k$ in the clockwise direction. Thus, for each bin we regress to this local orientation $\Delta_k$.

As a result, our network architecture consists of an encoder and two sets of decoders. The encoder takes the whole image as input and produces an $H\times W$ grid of high-level descriptors via a convolutional network. The decoders then use the feature descriptors to make junction predictions.  
Each junction is represented by $\p_{ij} = \big(\x_{ij}, c_{ij}, \{\theta_{ijk}, c_{ijk}^{\theta}\}_{k=1}^K \big)$, where $\x_{ij}$ is the coordinates of the junction center, $c_{ij}\in [0,1]$ is the confidence score for the presence of a junction in the $ij$-th grid cell, $\theta_{ijk}$ is the angle for the branch in the $k$-th bin, and $c_{ijk}$ is the confidence score for the bin. The two sets of decoders predict the junction center and the branches respectively. Each FCN decoder is simply a convolutional layer followed by a regressor, as shown in Fig.~\ref{fig:system} top.

Unlike local junctions, the junctions we aim to detect each is formed by the intersection of two or more long line segments (the branches). While the junction detection does not explicitly rely on edge/line detection as an intermediate step, the knowledge about the associated edges is indirectly learned by enforcing the network to make correct detection of the branches and their directions.


\subsubsection{Loss Function}\vspace{-2mm}

To guide the learning process towards the desired output, our loss function consists of four modules. Given a set of ground truth junctions $\P = \{\p_1, \ldots, \p_N\}$ in an image, we write the loss function as follows:\vspace{-2mm}
\begin{equation}
L = \lambda_{conf}^c L_{conf}^c +  \lambda_{loc}^c L_{loc}^c +  \lambda_{conf}^b L_{conf}^b  +  \lambda_{loc}^b L_{loc}^b. 
\label{eq:obj}\vspace{-2mm}
\end{equation}
In the following, we explain each term in more detail.


\noindent{\bf Junction center confidence loss $L_{conf}^c$.} The junction center confidence decoder predicts a score $\hat{c}_{ij}$, indicating the probability of a junction for each grid cell. Let ${c}_{ij}$ be the ground truth binary class label, we use the cross-entropy loss:
\begin{equation}\label{eq:loss1}
\small
L_{conf}^c = - \frac{1}{H\times W} \sum_{i, j} E(\hat{c}_{ij}, {c}_{ij}).
\end{equation}

\noindent{\bf Junction center location loss $L_{loc}^c$.} The junction center location decoder predicts the relative position $\hat{\x}_{ij}$ of a junction for each grid cell. We compare the prediction with each ground truth junction using the $\ell_2$ loss:\vspace{-2mm}
\begin{equation}
\small
L_{loc}^c = - \frac{1}{N}\sum_{n=1}^N \| \hat{\x}_{f(n)} - \x_{f(n)}\|_2^2,\vspace{-2mm}
\end{equation}
where $f(n)$ returns the index of the grid cell that the $n$-th ground truth junction falls into, and ${\x}_{f(n)}$ is the relative position of the ground truth junction w.r.t. that cell center.


\noindent{\bf Junction branch confidence loss $L_{conf}^b$.} The junction branch confidence decoder predicts a score $\hat{c}_{ijk}^{\theta}$ for each bin in each grid cell, indicating the probability of a junction branch in that bin. Similar to the junction center confidence loss above, we use the cross-entropy loss to compare the predictions with the ground truth labels. The only difference is that we only consider those grid cells in which a ground truth junction is present:\vspace{-2mm}
\begin{equation}\label{eq:loss2}
\small
L_{conf}^{b} = - \frac{1}{N\times K} \sum_{n=1}^N \sum_{k=1}^K E(\hat{c}_{f(n),k}^{\theta}, {c}_{f(n),k}^{\theta}).\vspace{-2mm}
\end{equation}
   
\noindent{\bf Junction branch location loss $L_{loc}^b$.} Similar to the junction center location loss, we first decide, for each ground truth junction, the indices of the bins that its branches fall into, denoted as $g(r), r=1, \ldots, R_n$, where $R_n$ is the order of $\p_n$. Then, we compare our predictions with the ground truth using the $\ell_2$ loss:\vspace{-2mm}
\begin{small}
\begin{equation}
L_{loc}^b  = - \frac{1}{N}\sum_{n=1}^N \big( \frac{1}{R_n}\sum_{r=1}^{R_n} \|\hat{\theta}_{f(n), g(r)} - {\theta}_{f(n), g(r)}\|_2^2  \big).\vspace{-2mm}
\end{equation}
\end{small}


\noindent{\bf Implementation details.} We construct our model to encode an image into $60\times 60$ grid of 256-dimensional features. Each cell in the grid is responsible for predicting if a junction is present in the corresponding image region. Our encoder is based on Google's Inception-v2 model~\cite{Szegedy2016RethinkingTI}, which extracts multi-scale features and is well-suited for our problem. For our problem, we only use the early layers in the Inception network, i.e., the first layer to ``Mixed\_3b''. Each decoder consists of a $3\times3\times256$ convolutional layer, followed by a ReLU layer and a regressor. Note that the regressor is conveniently implemented as $1\times1\times d$ convolutional layer, where $d$ is the dimension of the output.

The default values for the weights in Eq.~\eqref{eq:obj} are set to the following: $\lambda_{conf}^c = \lambda_{conf}^b = 1, \lambda_{loc}^c=\lambda_{loc}^b=0.1$. We choose the number of bins $K=15$. Our network is trained from scratch with the Stochastic Gradient Descent (SGD) method. The momentum parameter is set to 0.9, and the batch size is set to 1. We follow the standard practice in training deep neural networks to augment the data with image domain operations including mirroring, flipping upside-down, and cropping. The initial learning rate is set to 0.01. We decrease it by a multiple of 0.1 after every 100,000 iterations. Convergence is reached at 300,000 iterations.




\subsection{Line Detection}
Next we design and train a convolutional neural network (Fig.~\ref{fig:system} bottom) to infer line information from RGB images. The network predicts for each pixel $p$ whether it falls on a (long) line $\l$. To suppress local edges, short lines, and curves, the predicted value $h(p)$ (of the heat map) at pixel $p$ is set to be the length of the line it belongs to. Given an image with ground truth lines $\L$, the target value for $h(p)$ is defined to be:\vspace{-2mm}
\begin{equation}
    h(p) = 
    \begin{cases} 
        d(\l)     &   \text{$p$ is on a line $\l$ in $\L$}, \\
        0       &   \text{$p$ is not on any line in $\L$}, 
    \end{cases} \vspace{-2mm}
\end{equation}
where $d(\l)$ is the length of the line $\l$. Let $\hat{h}(p)$ be the estimated heatmap value, then the loss function we try to minimize the $\ell_2$ loss:\vspace{-2mm}
\begin{equation}
L = \sum_{i,j} \| \hat{h}(p_{ij}) - h(p_{ij})\|^2_2.\vspace{-2mm}
\end{equation}
where the sum is over all pixels of the image. 

\noindent{\bf Implementation details.}  The network architecture is inspired by the Stacked Hourglass network~\cite{NewellYD16}. It takes a $320 \times 320 \times 3$ RGB image as input, extracts a $80 \times 80 \times 256$ feature maps via three Pyramid Residual Modules (PRM), see Fig. \ref{fig:system} bottom. The feature maps then go through five stacked hourglass modules, followed by two fully convolutional and ReLU layers ($5\times 5\times 32$ and $5\times 5\times 16$) and a $5 \times 5 \times 1$ convolutional layer  to output a $320 \times 320 \times 1$ pixel-wise heat map. The detailed pyramid residual module and stacked hourglass module can be found in ~\cite{NewellYD16}.

During training, we adopt the Stochastic Gradient Descent (SGD) method. The momentum parameter is set to 0.9, and the batch size is set to 4. Again, we augment the data with image domain operations including mirroring and flipping upside-down. The initial learning rate is set to 0.001. We decrease it by a multiple of 0.1 after 100 epochs. Convergence is reached at 120 epochs.

Notice that we have used an Inception network for junction detection whereas an hourglass network for line detection. In junction detection, we are not interested in the entire support of the line, hence the receptive field of an Inception network is adequate for such tasks. However, we find that for accurately detecting lines with large spatial support, the Stacked Hourglass network works much better due to its large (effective) receptive field. In addition, our experiment also shows that above length-dependent $\ell_2$ loss is more effective than the cross-entropy cost often used in learning-based edge detection.

\subsection{Combine Junctions and Lines for Wireframe}
The final step of the system is to combine the results from junction detection and line detection to generate a wireframe $\W$ for the image, which, as mentioned before, consists of a set of junction points $\P$ connected by a set of line segments $\L$. 

Specifically, given a set of detected junctions $\{\p_i\}_{i=1}^N$ and a line heat map $h$, we first apply a threshold $w$ to convert $h$ into a binary map $\mathcal{M}$. Then, we construct the wireframe based on the following rules and procedure:\vspace{-1mm}
\begin{enumerate}
\item The set $\P$ is initialized with the output from the junction detector. A pair of detected junctions $\p$ and $\mathbf{q} \in \P$ are connected by a line segment $\l = (\p, \mathbf{q})$ if they are on (or close to be on) each other's branches, and we add this segment $\l$ to $\L$. If there are multiple detected junctions on the same branch of a junction point $\p$, we only keep the shortest segment to avoid overlap.\footnote{Hence we are less interested in detecting a straight line with the longest possible support, instead, we are interested in its incidence relationship with other lines and junctions.} \vspace{-1mm}
\item For any branch of a junction $\p$ that is not connected to any other junction, we attempt to recover additional line segment using $\mathcal{M}$. We first find the farthest line pixel $\mathbf{q}_{\mathcal{M}}$ (pixel $p$ is a line pixel if $\mathcal{M}(p) = 1$) that is also on the ray starting at $\p$ along the branch. Then, we find all the intersection points $\{\q_1, \ldots, \q_S\}$ of line segment $(\p, \q_{\mathcal{M}})$ with existing segments in $\L$. Let $\q_0 = \p_i$ and $\q_{S+1} = \q_{\mathcal{M}}$, we calculate the line support ratio $\kappa(\q_{s-1},\q_s), s=\{1, \ldots, S, S+1\}$, for each segment. Here, $\kappa$ is defined as the ratio of the number of line pixels to the total length of the segment. If $\kappa$ is above a threshold, say $0.6$, we add the segment to $\L$ and its endpoints to $\P$.
\end{enumerate}
Notice that both the sets $\P$ and $\L$ may have {\em two} sources of candidates. For the junction set $\P$, besides those directly detected by the junction detection, the line segments could also produce new intersections or endpoints that were missed by the junction detection. For the line segment set $\L$,  it could come from branches of the detected junctions and the line detection. 

We leave more detailed description of the algorithm to the supplementary material. Of course, there could be more advanced ways to merge the detected junctions and line heat map which we will explore in future work. Nevertheless, from our experiments (see next section), we find that the results from junction detection and line detection are rather complementary to each other and the above simple procedure already produces rather decent results. 

\section{Experiments}

In this section, we conduct extensive experiments to evaluate the quality of junctions and final wireframes generated by our method, and compare it to the state-of-the-art. All experiments are conducted on one NVIDIA Titan X GPU device. In testing phase, our method runs at about two frames per second, thus our method is potentially suitable for applications which require real-time processing.

\subsection{Datasets and Evaluation Metrics}

For performance evaluation, we split our wireframe dataset into a training set and a testing set. Among the 5,462 images in the dataset, 5,000 images are randomly selected for training and validation, and the remaining 462 images are used for testing. For junction detection (Section~\ref{sec:exp:junction}), we compare the junctions detected by any method with the ground truth junctions (Fig.~\ref{fig:data}, second row). For wireframe construction (Section~\ref{sec:exp:wireframe}), we compare the line segments detected by any method with the ground truth line segments labeled by human subjects (Fig.~\ref{fig:data}, first row).

For both junction detection and wireframe construction experiments, all methods are the evaluated quantitatively by means of the \emph{recall} and \emph{precision} as described in~\cite{MartinFM04, MaireAFM08, XiaDG14}. In the context of junction detection, recall is the fraction of true junctions that are detected, whereas precision is the fraction of junction detections that are indeed true positives. In the context of wireframe construction, recall is the fraction of line segment pixels that are detected, whereas precision is the fraction of line segment pixels that are indeed true positives.

Specifically, let $G$ denote the set of ground truth junctions (or line segment pixels), and $Q$ denote the set of junctions (or line segment pixels) detected by any method, the precision and recall are defined as follows:\vspace{-2mm}
\begin{equation}
\mbox{Precision} \doteq {\left| G\cap Q\right|}/{\left| Q \right|}, \;\; \mbox{Recall} \doteq {\left| G\cap Q\right|}/{\left| G \right|}.\vspace{-2mm}
\end{equation}
Note that, following the protocols of previous work~\cite{MartinFM04, MaireAFM08, XiaDG14},  the particular measures of recall and precision allow for some small tolerance in the localization of the junctions (or line segment pixels). In this paper, we set the tolerance to be 0.01 of the image diagonal.


\subsection{Junction Detection Comparison}
\label{sec:exp:junction}
We compare our junction detection method with two recent methods, namely Manhattan junction detection (MJ)~\cite{RamalingamPJT13} and \emph{a contrario} junction detection (ACJ)~\cite{XiaDG14}.

\noindent{\bf MJ~\cite{RamalingamPJT13}}: This method detects Manhattan junctions formed by line segments in three principal orthogonal directions using a simple voting-based scheme. 
As the authors did not release their code, we use our own implementation of the method. Line segments are first detected using LSD~\cite{von2012lsd}, and then clustered using J-Linkage~\cite{Tardif09} to obtain the vanishing points. Note that this method only applies to scenes that satisfy the Manhattan world assumption. For fair comparison, we only keep the images in which three principal vanishing points are detected. An important parameter in our implementation is the maximum distance $d_{\max}$ between a line segment and a point $p$ for that line segment to vote for $p$. We vary the value $d_{\max} \in \{10, 20, 30, 50, 100, 200, 300, 500\}$ pixels.

\noindent{\bf ACJ~\cite{XiaDG14}}: This method relies on statistical modeling of image gradients and an \emph{a contrario} approach to detect junctions. Specifically, meaningful junctions are detected as those which are very unlikely under a null hypothesis $\mathcal{H}_0$, which is defined based on the distribution of gradients of arbitrary natural images. In the method, each candidate junction is associated with a strength value depending on the image gradients around it. Then, the candidate junction is validated with a threshold, which is derived by controlling the number of false detections, $\epsilon$, in an image following $\mathcal{H}_0$. For the experiments, we use the implementation provided by the authors of~\cite{XiaDG14} and vary the value $\epsilon \in \{10^{-3}, 10^{-2}, 10^{-1}, 1, 10^1, 10^2, 10^3\}$.

\begin{figure}[t]
    \centering
    \includegraphics[width = 1.9in]{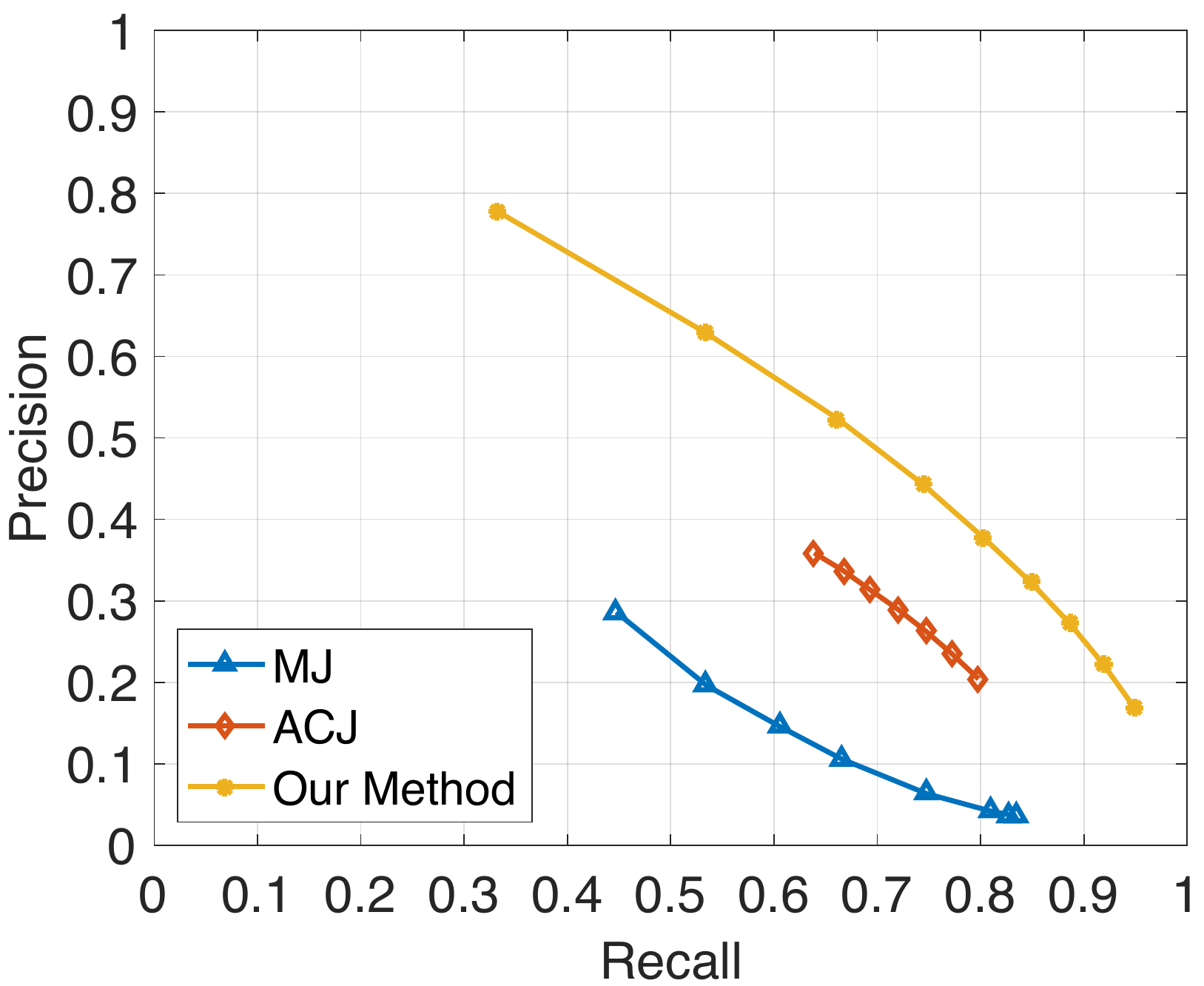}
    \caption{The precision-recall curves of different junction detection methods on our test dataset.}\vspace{-3mm}
    \label{fig:pr}
\end{figure}

\begin{figure*}[t!]
\centering

\includegraphics[height=0.9in]{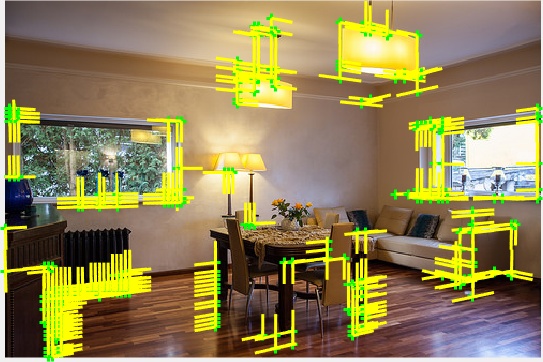}
\includegraphics[height=0.9in]{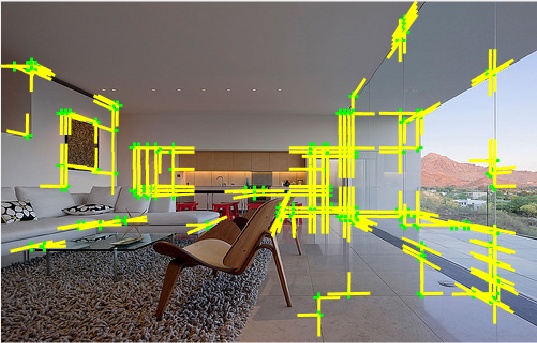}
\includegraphics[height=0.9in]{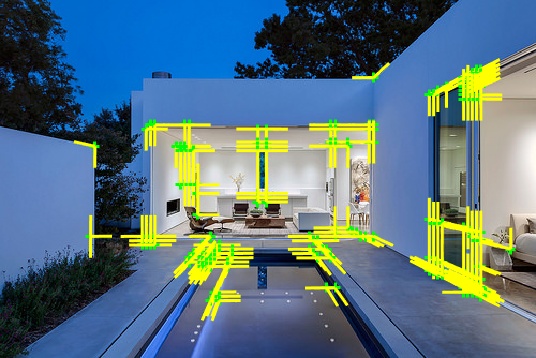}
\includegraphics[height=0.9in]{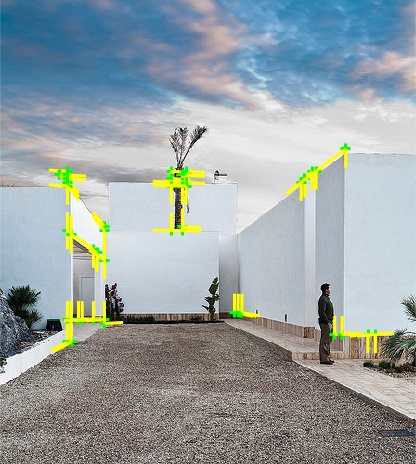}
\includegraphics[height=0.9in]{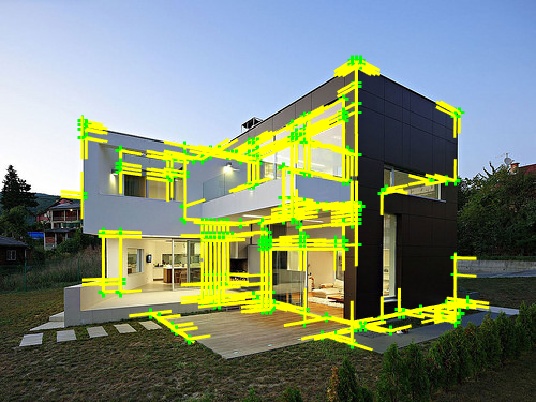}
\\
\includegraphics[height=0.9in]{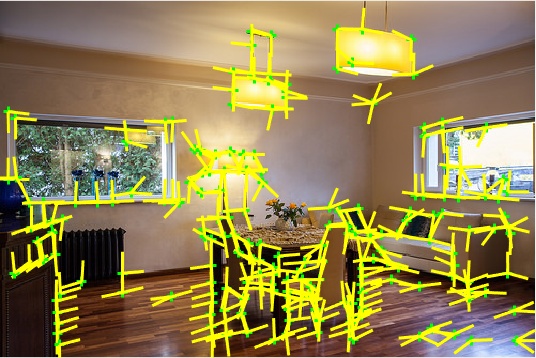}
\includegraphics[height=0.9in]{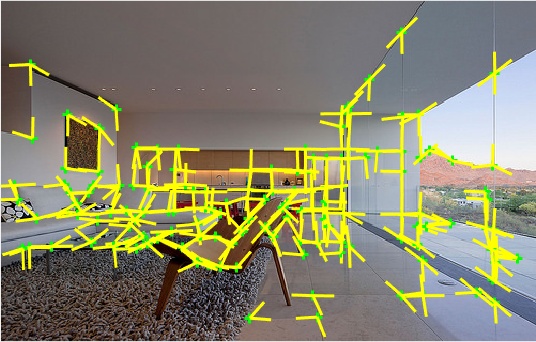}
\includegraphics[height=0.9in]{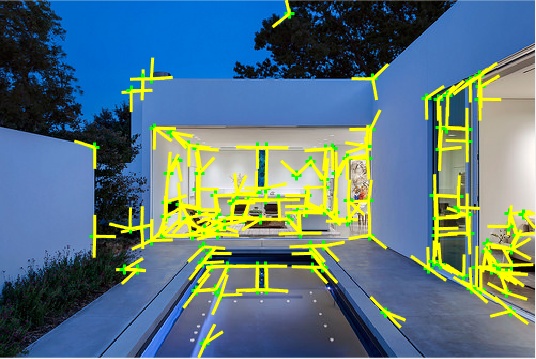}
\includegraphics[height=0.9in]{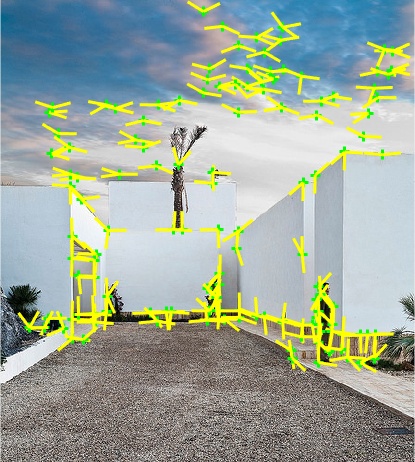}
\includegraphics[height=0.9in]{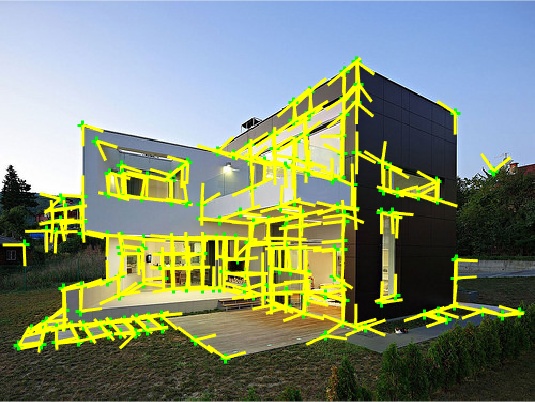}
\\
\includegraphics[height=0.9in]{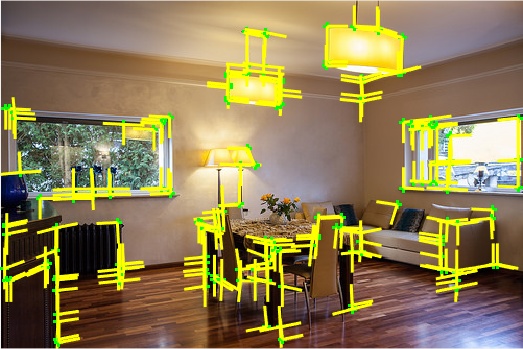}
\includegraphics[height=0.9in]{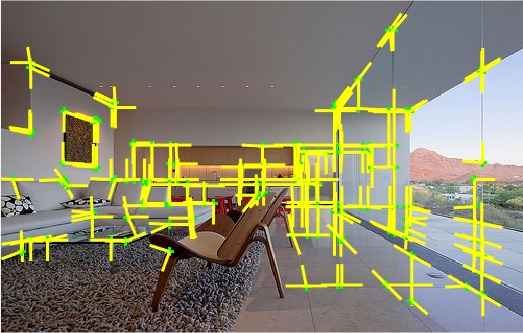}
\includegraphics[height=0.9in]{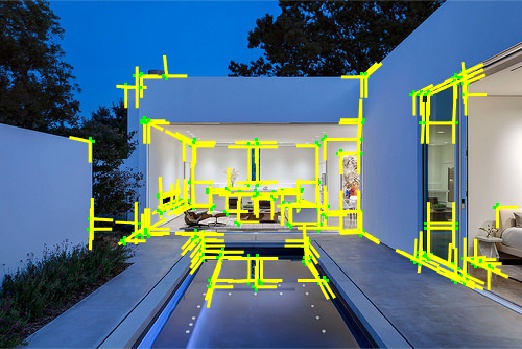}
\includegraphics[height=0.9in]{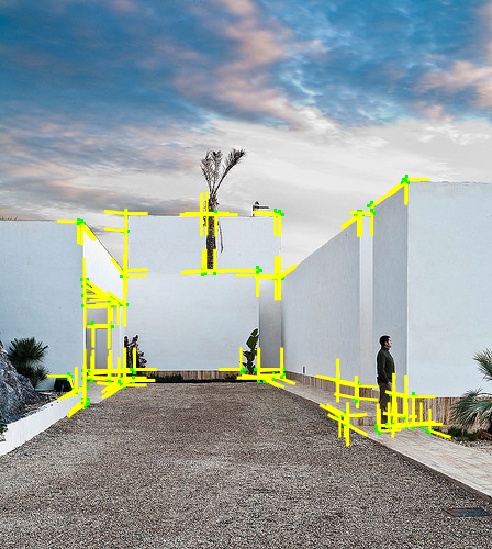}
\includegraphics[height=0.9in]{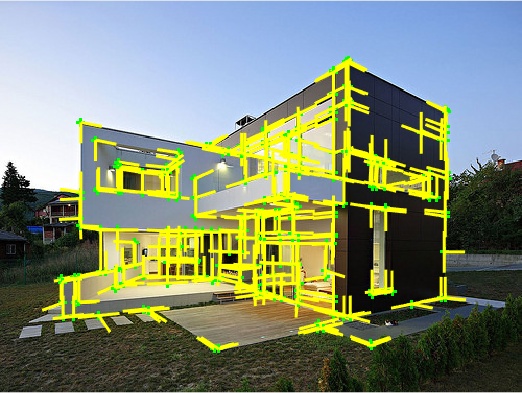}
\\
\caption{Junction detection results. {\bf First row:} MJ ($d_{\max}=20$). {\bf Second row:} ACJ ($\epsilon=1$). {\bf Third row:} Our method ($\tau=0.5$). } \vspace{-1mm}
\label{fig:results}
\end{figure*}

\noindent{\bf Performance comparison.} Fig.~\ref{fig:pr} shows the precision-recall curves of all methods on our new dataset. For our method, we vary the junction confidence threshold $\tau$ from 0.1 to 0.9. As one can see, our method outperforms the other methods by a large margin. Fig.~\ref{fig:results} compares qualitatively the results of all methods on our test data. Compared to the other two methods, MJ tends to miss important junctions due to the imperfect line segment detection results. Moreover, since MJ relies on local image features, it noticeably produces quite a few repetitive detections around some junctions. By directly modeling the image gradients, ACJ is able to find most junctions on the scene structures. However, as a local method, ACJ makes a lot of false predictions on textured regions (e.g., floor of the first, sky of the fourth image). In contrast, our method is able to detect most junctions intersected by salient lines, while minimizing the number of false detections. This is no surprise because our supervised framework implicitly encodes high-level structural and semantic information of the scene as it learns from the labeled data provided by humans.

\subsection{Line Segment Detection Comparison}
\label{sec:exp:wireframe}
In this section, we compare the wireframe results of our method with two state-of-the-art line segment detection methods, namely the Line Segment Detector (LSD) ~\cite{von2012lsd} and the Markov Chain Marginal Line Segment Detector (MCMLSD)~\cite{Almazan17}. We test and compare with these methods on both our new dataset and the York Urban dataset \cite{Denis2008EfficientEM} used in the work of MCMLSD \cite{Almazan17}.

\noindent{\bf LSD~\cite{von2012lsd}}: This method is a linear-time line segment detector that requires no parameter tuning. It also uses an \emph{a contrario} approach to control the number of false detections. In this experiment, we use the code released by the authors\footnote{http://www.ipol.im/pub/art/2012/gjmr-lsd/} and vary the threshold for $-\log$(NFA) (NFA is the number of false alarms) in $0.01\times \{1.75^0, 1.75^1, 1.75^2, ..., 1.75^{19}\}$.

\noindent{\bf MCMLSD~\cite{Almazan17}}: This method proposes a two-stage algorithm to find line segments. In the first stage, it uses the probabilistic Hough transform~\cite{matas2000robust} to identify globally optimal lines. In the second stage, it searches each of these lines for their supports (segments) in the image, which can be modeled as labeling hidden states in a linear Markov chain. In this experiment, we use the code released by the authors.\footnote{http://www.elderlab.yorku.ca/resources/} Be aware that authors of \cite{Almazan17} have introduced a different metric than ours that tends to penalize over-segmentation. Hence our metric can be unfair to their method. Nevertheless, our metric is more appropriate for wireframe detection as we prefer to interpret a long line as several segments between junctions if it intersects with other lines.

\begin{figure}[t]
    \centering
    \includegraphics[width = 1.62in]{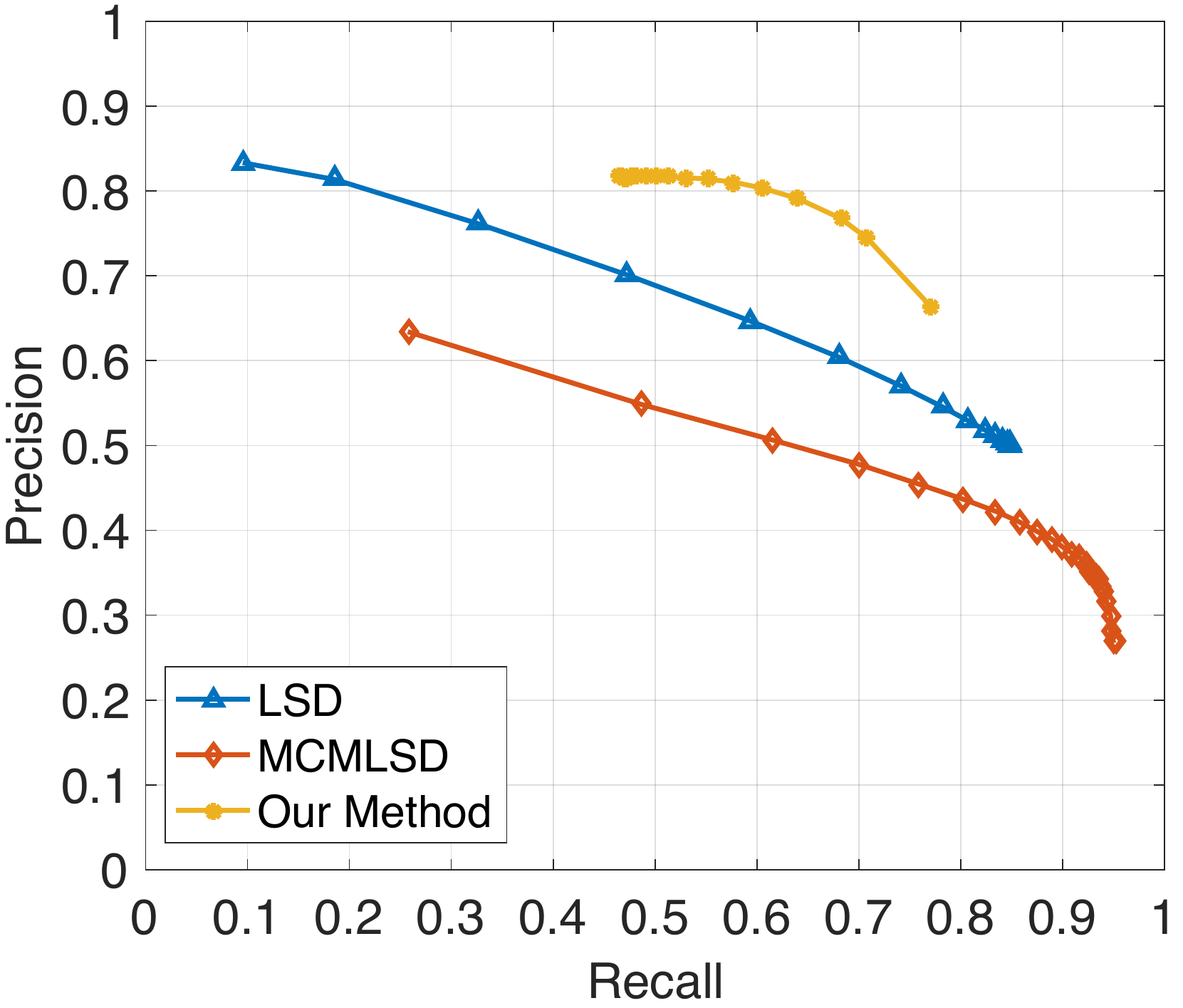}
    \includegraphics[width = 1.62in]{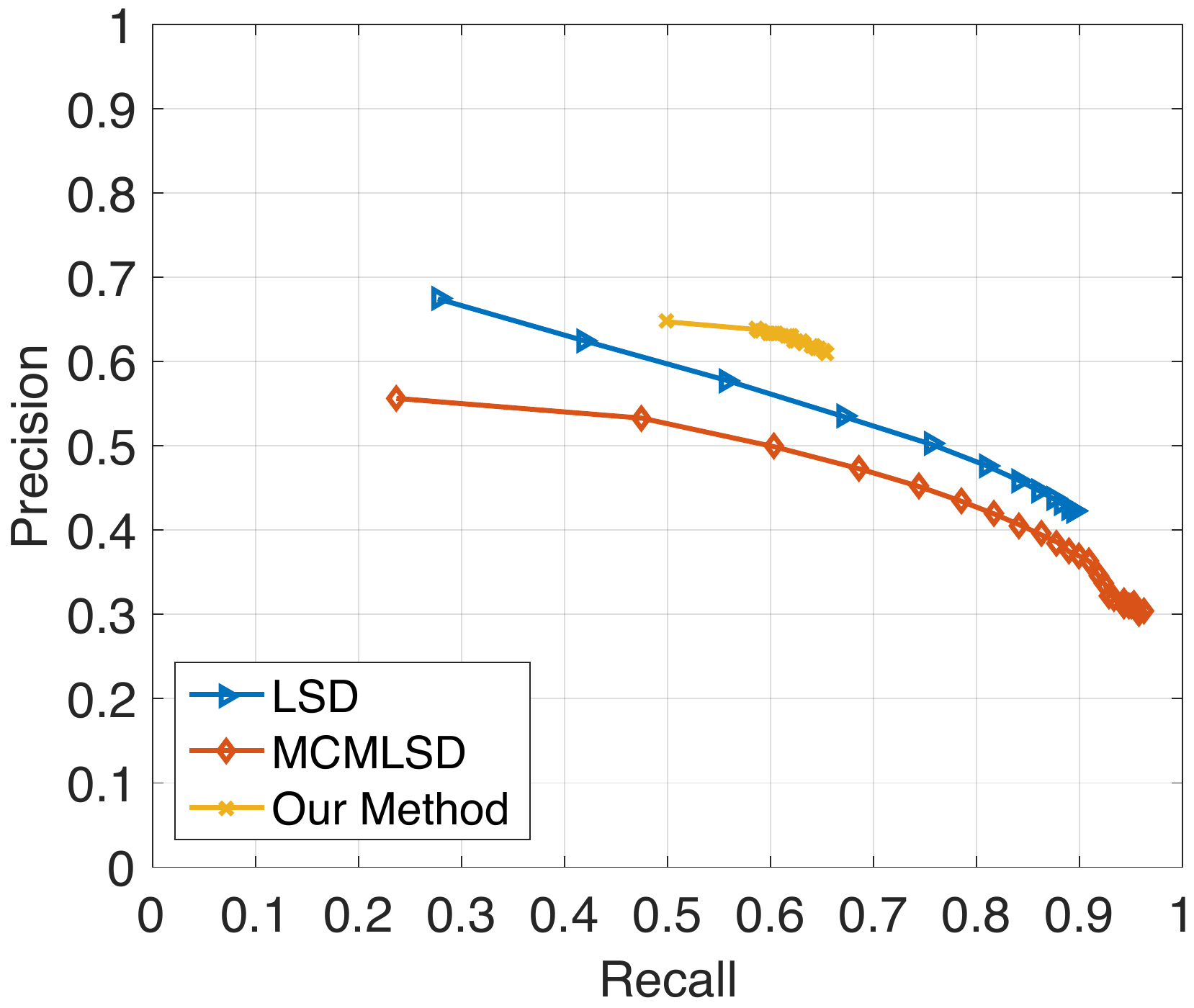}
    \caption{The precision-recall curves of different line segment detection methods. {\bf Left:} on our test dataset. {\bf Right:} on the York Urban dataset \cite{Denis2008EfficientEM}.} \vspace{-3mm}
    \label{fig:LinePR-ours} \label{fig:LinePR-york}
\end{figure}

\begin{figure*}[t!]
\centering
\includegraphics[height=0.88in]{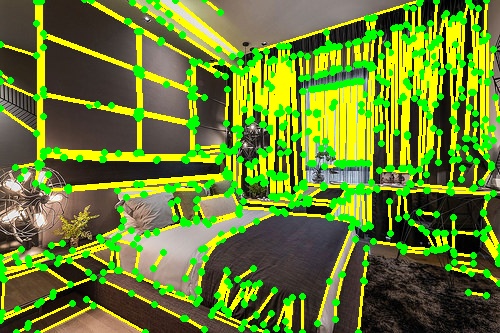}
\includegraphics[height=0.88in]{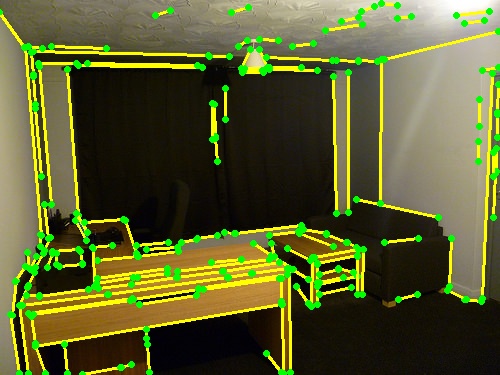}
\includegraphics[height=0.88in]{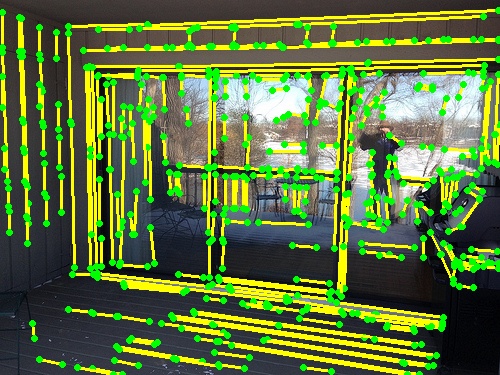}
\includegraphics[height=0.88in]{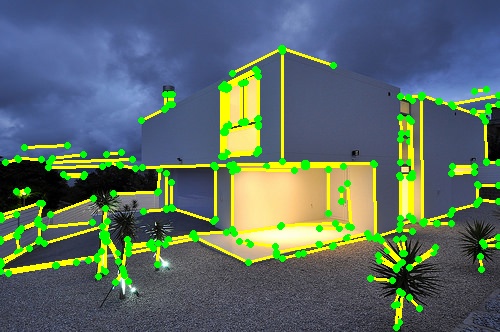}
\includegraphics[height=0.88in]{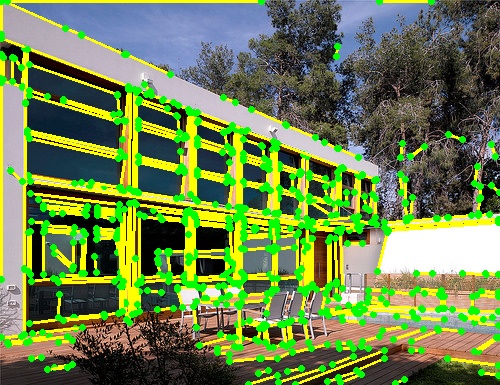}
\\
\includegraphics[height=0.88in]{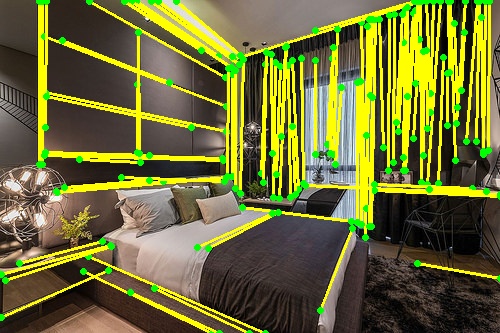}
\includegraphics[height=0.88in]{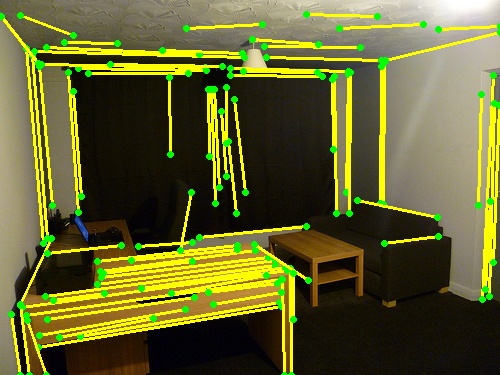}
\includegraphics[height=0.88in]{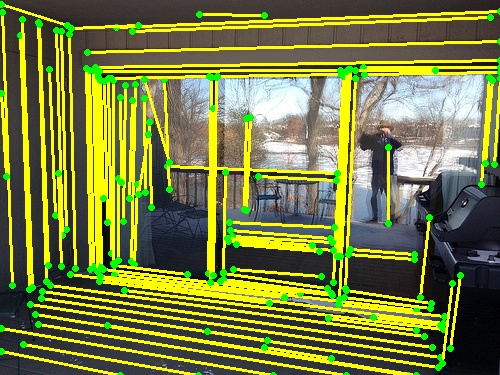}
\includegraphics[height=0.88in]{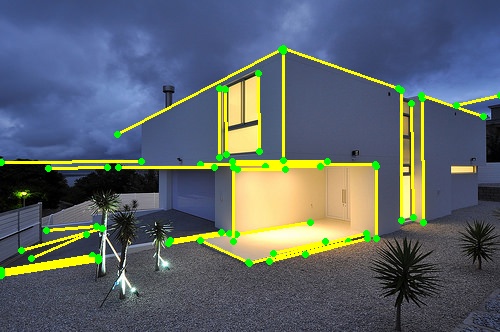}
\includegraphics[height=0.88in]{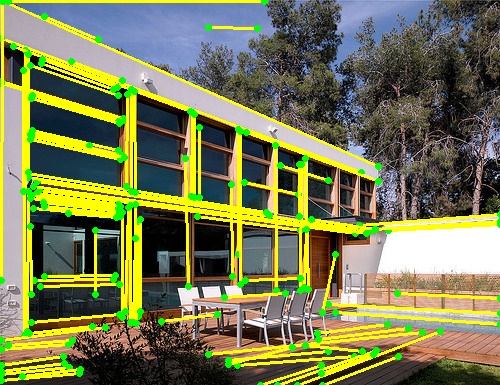}
\\
\includegraphics[height=0.88in]{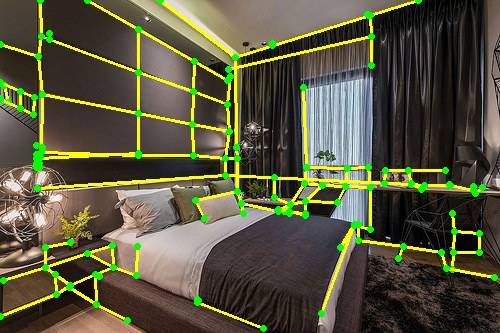}
\includegraphics[height=0.88in]{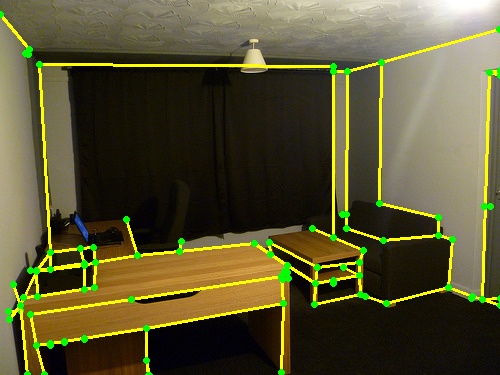}
\includegraphics[height=0.88in]{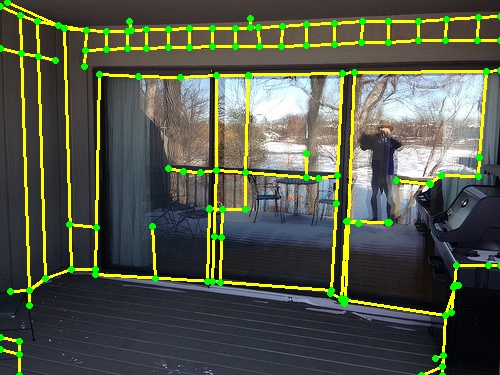}
\includegraphics[height=0.88in]{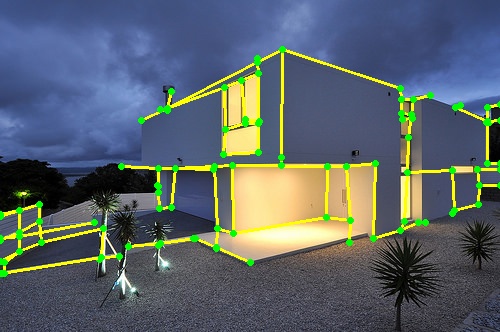}
\includegraphics[height=0.88in]{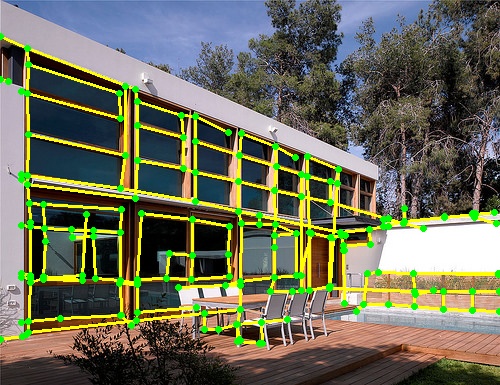}
\\
\includegraphics[height=0.88in]{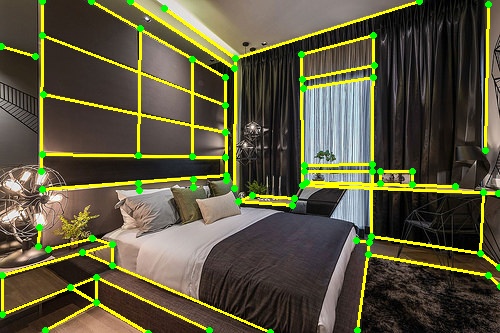}
\includegraphics[height=0.88in]{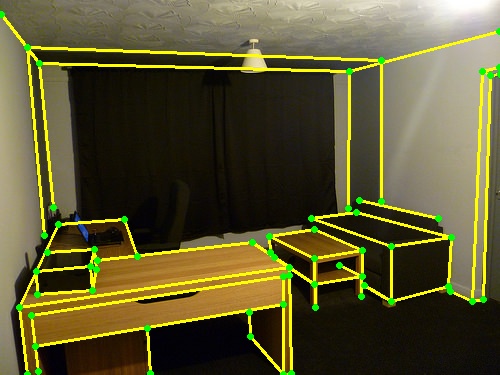}
\includegraphics[height=0.88in]{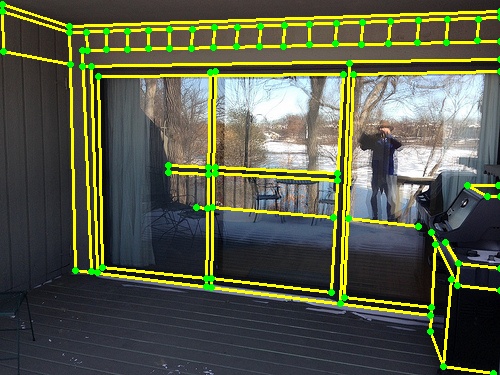}
\includegraphics[height=0.88in]{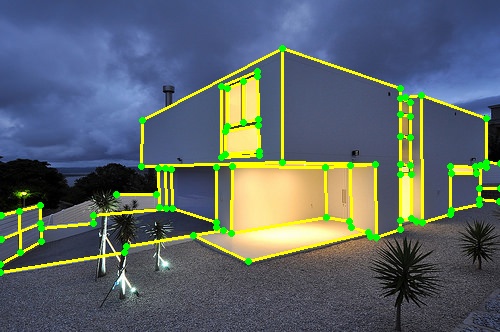}
\includegraphics[height=0.88in]{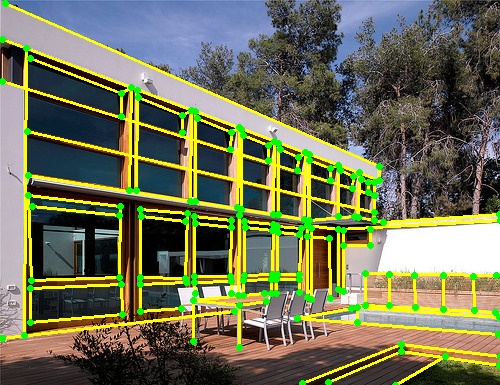}
\\
\caption{Line/wireframe detection results. {\bf First row:} LSD (-$\log$(NFA)  $> 0.01\times1.75^8$). {\bf Second row:} MCMLSD (confidence top 100). {\bf Third row:} Our method (line heat map $h(p) > 10$). {\bf Fourth row:} Ground truth.}\vspace{-4mm}
\label{fig:line-results}
\end{figure*}

\noindent{\bf Performance comparison.} Fig.~\ref{fig:LinePR-ours} shows the precision-recall curves of all methods on our dataset and the York Urban dataset, respectively. As one can see, our method outperforms the other methods by a significant margin on our dataset. The margin on the York Urban dataset is decent but not so large. According to \cite{Almazan17}, the labeling of the York Urban dataset is not as complete for salient line segments, hence it is not entirely suitable for the wireframe detection task here.  Fig.~\ref{fig:line-results} compares qualitatively the results of all methods on our test data. Since the other two methods rely on local measurements, they tend to produce many line segments on textured regions (e.g. curtain of the first image) which do not correspond to scene structures.

\section{Conclusion}
This paper has demonstrated the feasibility of parsing wireframes in images of man-made environments. The proposed method is based on combining junctions and lines detected from respective neural networks trained on a new large-scale dataset. Both quantitatively and qualitatively, the results of our method approximately emulate those labelled by humans. The junctions and line segments in a wireframe and their incidence relationships encode rich and accurate large-scale geometry of the scene and shape of regular objects therein, in a highly compressive and efficient manner. Hence results of this work can significantly facilitate and benefit visual tasks such as feature correspondence, 3D reconstruction, vision-based mapping, localization, and navigation in man-made environments. 

\noindent\textbf{Acknowledgement:} The project is supported by NSFC (NO. 61502304) and Program of Shanghai Subject Chief Scientist (A type) (No. 15XD1502900).

\newpage
{\small
\bibliographystyle{ieee}
\bibliography{cvpr18-parsing}
}

\clearpage

\section*{Supplementary}
\section*{A. Wireframe Construction Algorithm Detail}

Given an image, our wireframe construction algorithm takes a set of junctions $\{ \p_{i}\}_{i = 1}^N$, $\p_{i} = \big(\x_{i}, \{\theta_{ik}\}_{k=1}^{K_i} \big)$, and a line heat map $h$ as input. Note that for the junctions and their branches predicted by our network, we only keep those with confidence scores higher than certain thresholds $\tau_c$ and $\tau_b$, respectively. As a pre-processing step, we further adopt a strategy similar to non-maximum suppression to remove duplicate detections. 

\begin{algorithm}
	\caption{Wireframe Construction}
	\label{alg:construct}
	\begin{algorithmic}[1]
		\Require Junctions $\{ \p_{i}\}_{i = 1}^N$, $\p_{i} = \big(\x_{i}, \{\theta_{ik}\}_{k=1}^{K_i} \big)$, and a line heat map $h(p)$		\Ensure Wireframe $\W$ consisting of a set of junction points $\P$ connected by a set of line segments $\LL$
		\State Initialize $\P \leftarrow \O$, $\LL \leftarrow \O, \V \leftarrow \mathbf{0}$
		\State Binarize $h(p)$ with threshold $\omega$ into $\M$

		
		\For{$t_1 \in \{1, 2, \ldots, N_r\}$}
		\State $(i, k_1) \leftarrow \pi(t_1)$, $d_{\min} \leftarrow \infty$, $z \leftarrow 0$
		
		\For{$t_2 \in \{1, 2, \ldots, N_r\}$}
		\State $(j, k_2) \leftarrow \pi(t_2)$	
		

		
			\If{ $j \neq i$ \textbf{and} $\p_j$ on $\rr_{ik_1}$ \textbf{and} $\p_i$ on $\rr_{jk_2}$ }
				\If{$\|\x_i - \x_j\| < d_{\min}$}
					\State $d_{\min} \leftarrow \|\x_i - \x_j\|$, $z \leftarrow t_2$		
				\EndIf
			\EndIf
		\EndFor
		\If{$ z \neq 0$}
			\State $\V(t_1, z) \leftarrow 1$
		\EndIf
		
		\EndFor
		
		\ForAll{$t_1, t_2 \in \{1, 2, \ldots, N_r\}, t_1 \neq t_2$}
				\If{$\V(t_1, t_2) = 1$ \textbf{and} $\V(t_2, t_1) = 1$}
				\State $(i, k_1) \leftarrow \pi(t_1)$, $(j, k_2) \leftarrow \pi(t_2)$
				\State $\P \leftarrow \P \bigcup \{\p_i, \p_j\}$, $\LL \leftarrow \LL \bigcup \{(\p_i, \p_j)\}$
				\EndIf
				\EndFor
		
		\ForAll{$\rr_{ik}$ not matched to another ray}
		\State Find the intersection of $\rr_{ik}$ and image boundary $\q_b$
		\If{$\|x_i - \q_b\| \leq 0.05 \times m$}
			\State $\P \leftarrow \P \bigcup \{\p_i, \q_b\}$, $\LL \leftarrow \LL \bigcup \{(\p_i, \q_b)\}$
		
		\Else
			\State Find the farthest point $\q_{\M}$ along $\rr_{ik}$ on $\M$
			\State Find all intersections $\{\q_1, \ldots, \q_S\}$ of $(\p_i, \q_{\M})$ with segments in $\LL$
				\State $\q_0 \leftarrow \p_i$, $\q_{S+1} \leftarrow \q_{\M}$		
				\For{$s \in \{1, 2, ..., S, S+1\}$}
					\If{$\kappa(\q_{s-1}, \q_s)> 0.6$}
						\State $\P \leftarrow \P \bigcup \{\q_{s-1}, \q_s\}$
						\State $\LL \leftarrow \LL \bigcup \{(\q_{s-1}, \q_s)\}$
					\EndIf		
				\EndFor
		\EndIf
		\EndFor
	\end{algorithmic}
\end{algorithm}

Our wireframe construction algorithm is presented in Alg.~\ref{alg:construct}. In the algorithm, we first apply a threshold $\omega$ to convert the line heat map $h(p)$ into a binary map $\M$ (line 2). Note that this threshold $\omega$ is varied to obtain the precision-recall curve in our experiments on wireframe construction. The algorithm then proceeds as follows:

{\em First}, we connect all pairs of junctions which are aligned with each other's branch directions (lines 3-22).
Let $\rr_{ik}$ represent the ray starting at $\p_i$ along its $k$-th branch. We collect all possible rays as $\mathcal{R} = \{\rr_{11}, ..., \rr_{1K_1}, ..., \rr_{i1}, ..., \rr_{iK_i}, ...\}$, and use $(i,k) = \pi(t)$ to map the $t$-th ray in $\mathcal{R}$ to its junction index $i$ and branch index $k$. Then, for the rays in $\mathcal{R}$, we use $\V \in \mathbb{R}^{N_r\times N_r}$, $N_r = |\mathcal{R}|$, to record the indices of the corresponding ray/branch of the closest opposite junction. Specifically, $\forall t_1 \in \{1, \ldots, N_r\}$, we set $\V(t_1, t_2)$ to 1 if and only if (i) $\p_i$ is the on the ray $\rr_{jk_2}$ and $\p_j$ is on the ray $\rr_{ik_1}$, where $(i,k_1)=\pi(t_1)$, $(j,k_2)=\pi(t_2)$, and (ii) the distance between $\p_i$ and $\p_j$ is the shortest among all such aligned pairs (lines 5-15). Then, we consider two rays are matched if $\V(t_1,t_2) = \V(t_2, t_1)=1$ and add the corresponding junctions and line segments to $\P$ and $\LL$, respectively (lines 17-21).

{\em Second}, for any ray $\rr_{ik}$ which fails to find a matching ray using the above procedure, we attempt to recover additional line segments using the line support $\M$ (lines 23-38). We consider the following cases:
\begin{itemize}
	\item[(a)] If the distance between $\p_i$ and $\q_b$, the intersection of $\rr_{ik}$ and the image boundary, is smaller than certain threshold (say $0.05 \times m$ where $m$ is the maximum of image width and height), we add $\{\p_i, \q_b\}$ and the connecting line segment to $\P$ and $\LL$, respectively (lines 24-26).
	\item[(b)] For a ray exceeding the length threshold in (a), we first find the farthest line pixel $\q_{\M}$ along the ray on $\M$. Then, we find all the intersection points $\{\q_1, \ldots, \q_S\}$ of line segment $(\p_i, \q_{\M})$ with existing segments in $\LL$ (lines 28-29). Let $\q_0 = \p_i$ and $\q_{S+1} = \q_{\M}$, we calculate the line support ratio $\kappa(\q_{s-1},\q_s), s=\{1, \ldots, S, S+1\}$, for each segment. Here, $\kappa$ is defined as the ratio of line pixels (pixel $p$ is a line pixel if $\M(p) = 1$) to the total length of the segment. If $\kappa$ is above a threshold, say $0.6$, we add the segment to $\LL$ and its endpoints to $\P$ (lines 30-36).	 
	\end{itemize}

\nop{
	\begin{figure}[t]
		\centering
		\begin{tabular}{cc}
			\hspace{-3mm}    \includegraphics[height = 1.25in]{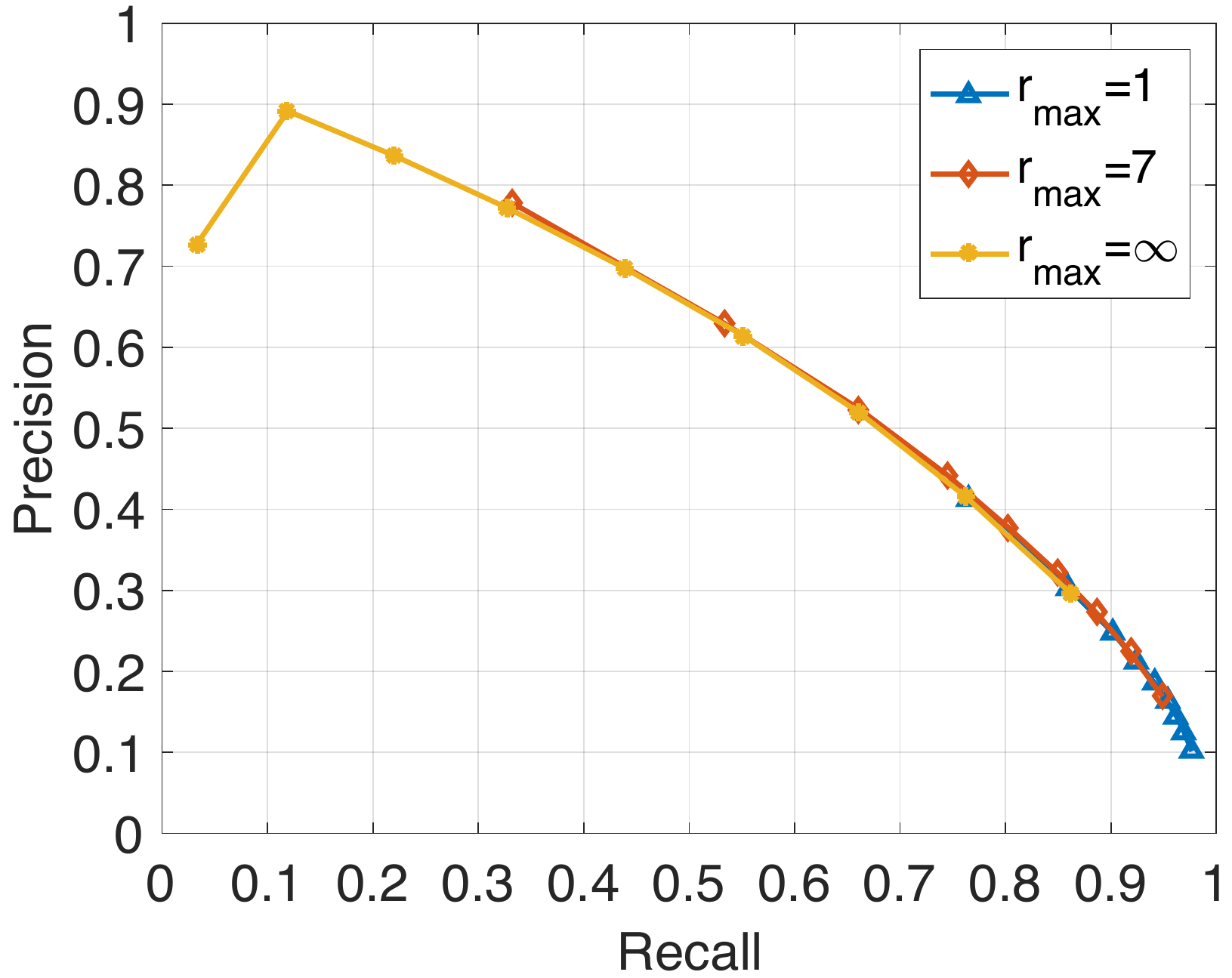} &
			\hspace{-3mm}    \includegraphics[height = 1.25in]{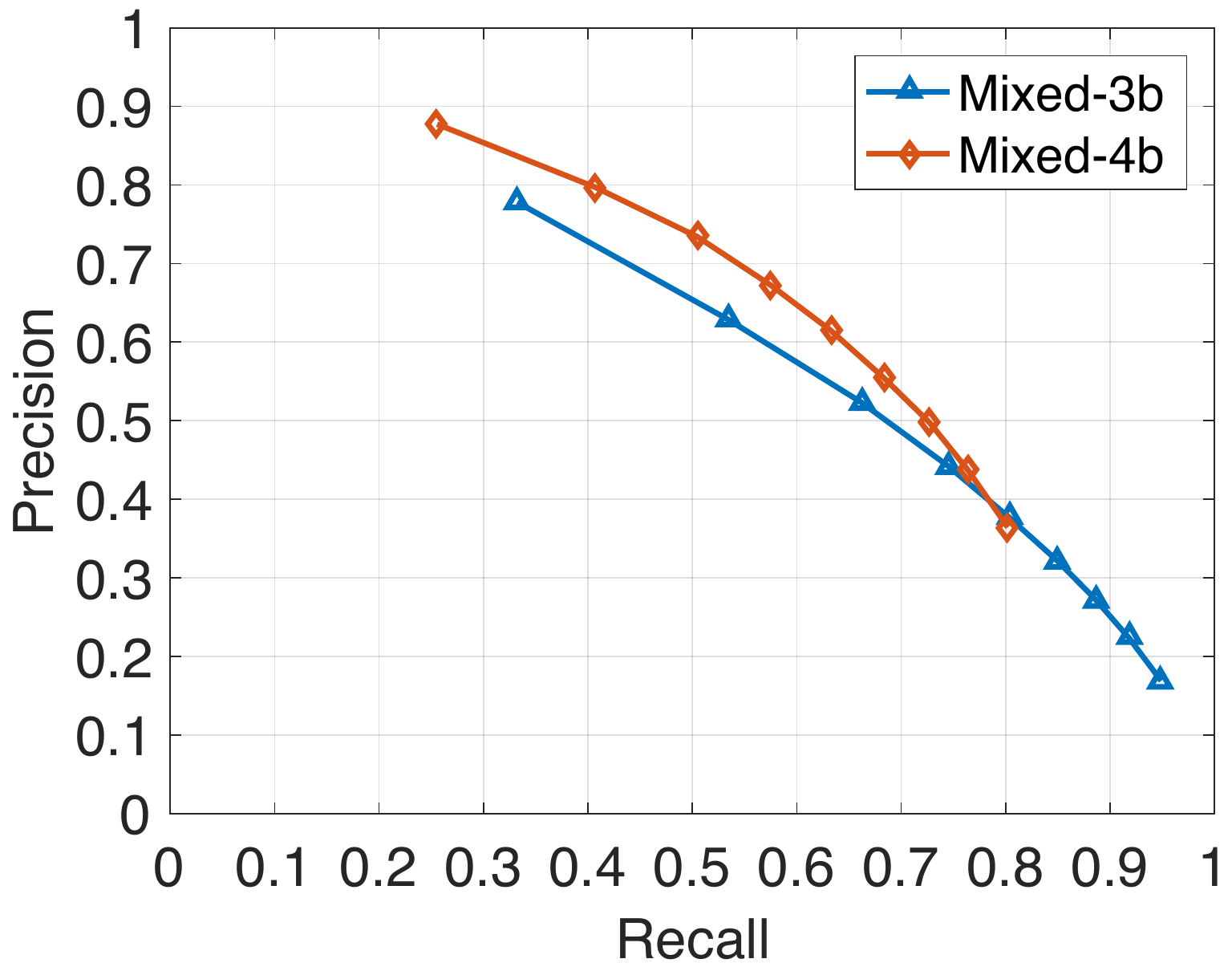} \\
			(a) $r_{\max}$ & (b) Network depth
		\end{tabular}
		\caption{Experiments on junction detection network parameters.}
		\label{fig:para}
	\end{figure}
}

\begin{figure}[t]
	\centering
	\begin{tabular}{c}
		\includegraphics[height = 1.8in]{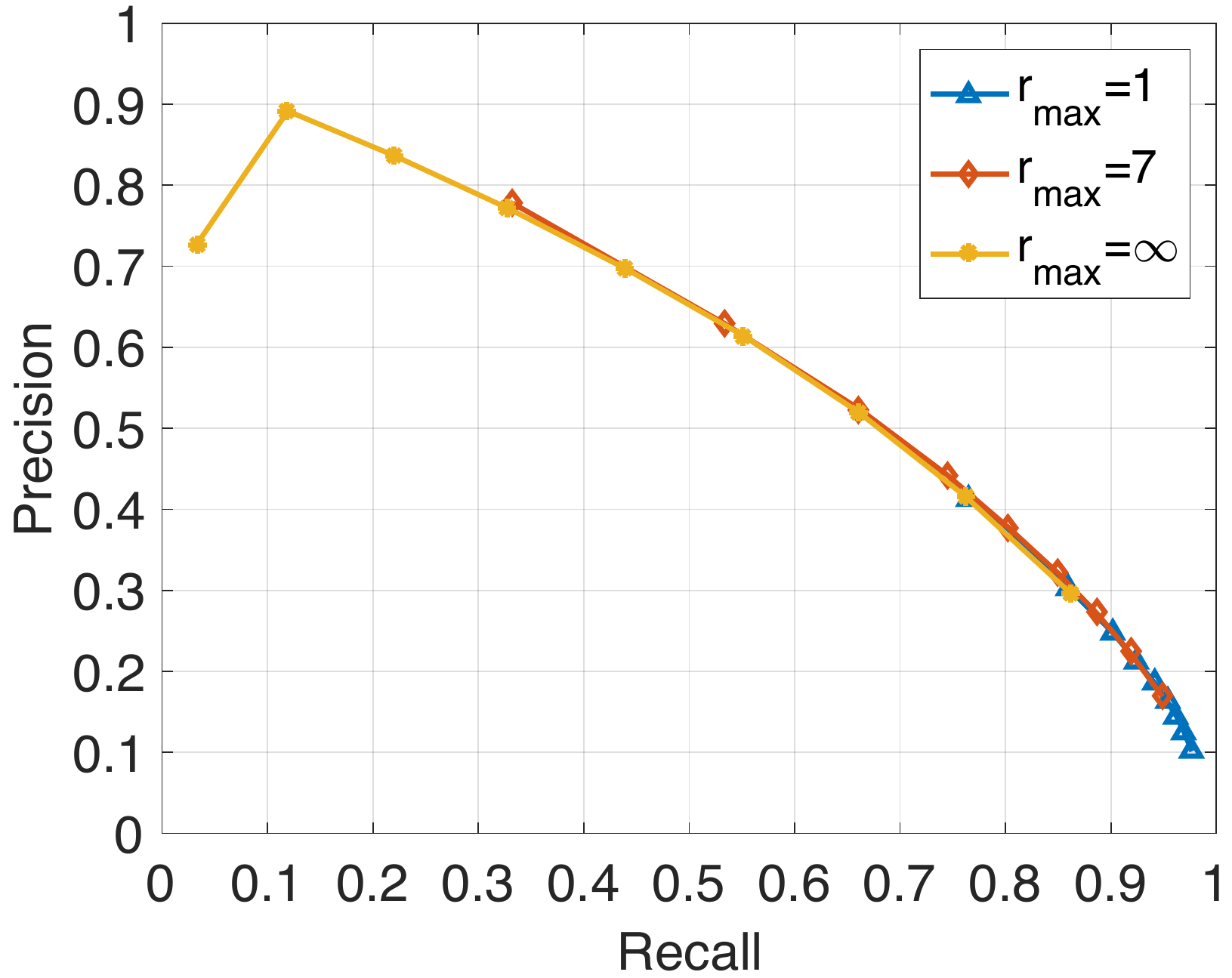} \\
		(a) $r_{\max}$\\
		\includegraphics[height = 1.8in]{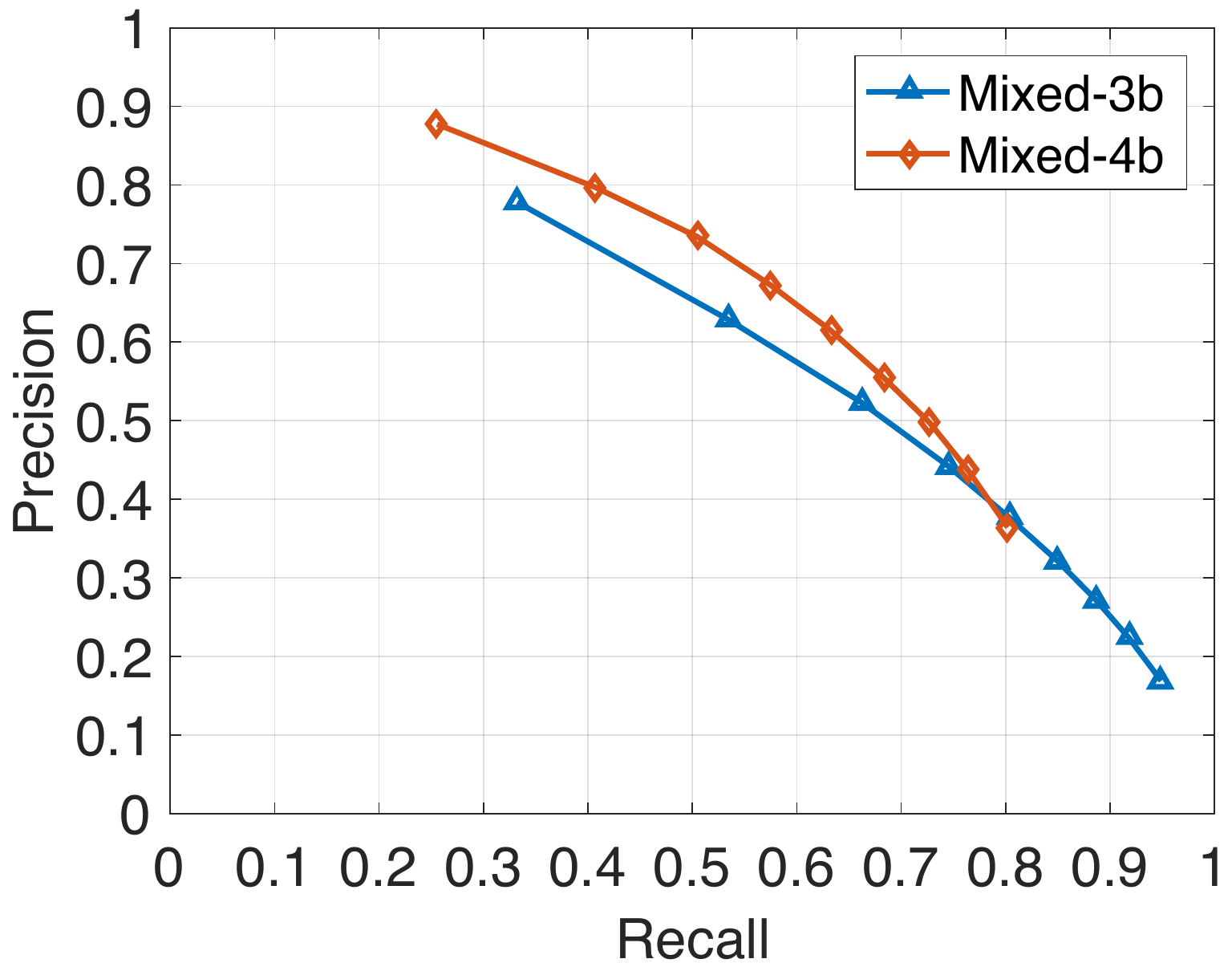} \\
		(b) Network depth
	\end{tabular}
	\caption{Experiment on junction detection network parameters.}
	\label{fig:para}
\end{figure}

\section*{B. Additional Experiments}

\subsection*{B.1. Experiment on Junction Detection Network Parameters}

In this section, we examine the choices of two important hyper-parameters in our junction detection network.

\medskip
\noindent{\bf Effect of balancing positive and negative samples.} In this experiment, we vary the value $r_{\max}$, which controls the maximum ratio between negative and positive samples at each iteration. Note that setting $r_{\max} = \infty$ is equivalent to using all grid cells during training. We can observe in Figure~\ref{fig:para}(a) that the precision-recall curves largely overlap. But as $r_{\max}$ increases, the curve shifts toward the high-precision-low-recall regime, and vice versa. For example, when $r_{\max} = 1$, the precision and recall at $\tau=0.5$ are 0.19 and 0.94, respectively. And when $r_{\max} = \infty$, the precision and recall at $\tau=0.5$ are 0.70 and 0.44, respectively. Note that this has an important implication in practice, as human annotators tend to miss true junctions much more often than labelling wrong junctions. Empirically, we have found that $r_{\max}=7$ yields more satisfactory results.

\medskip
\noindent{\bf Going deeper.} It is also interesting to investigate how the network depth of the encoder affects the performance. In this experiment, we compared two different choices based on Google Inception-v2, namely the first layer to ``Mixed\_3b'', and the first layer to ``Mixed\_4b''. Note that the latter has a larger depth and receptive field, at the cost of spatial resolution ($30\times 30$). As one can see in Figure~\ref{fig:para}(b), increasing the depth (i.e., predicting at the ``coarser'' level) results in higher precision but lower recall. This suggests possibilities to further improve the performance of our method using a ``skip-net'' architecture, that is, combining predictions at multiple levels. We leave this for future work.


\subsection*{B.2. Experiment on Line Segment Detection}

In this experiment, we study the possibility of extracting line segments \emph{directly} from the pixel-wise line heat map predicted by our network (i.e., without using junctions). To this end, we simply perform a probabilistic hough transform~\cite{matas2000robust} on the line heat map to generate line segments. We compare the results with LSD, MCMLSD, and our full wireframe construction method. 

\begin{figure}[t]
	\centering
	\begin{tabular}{c}
		\includegraphics[height = 1.8in]{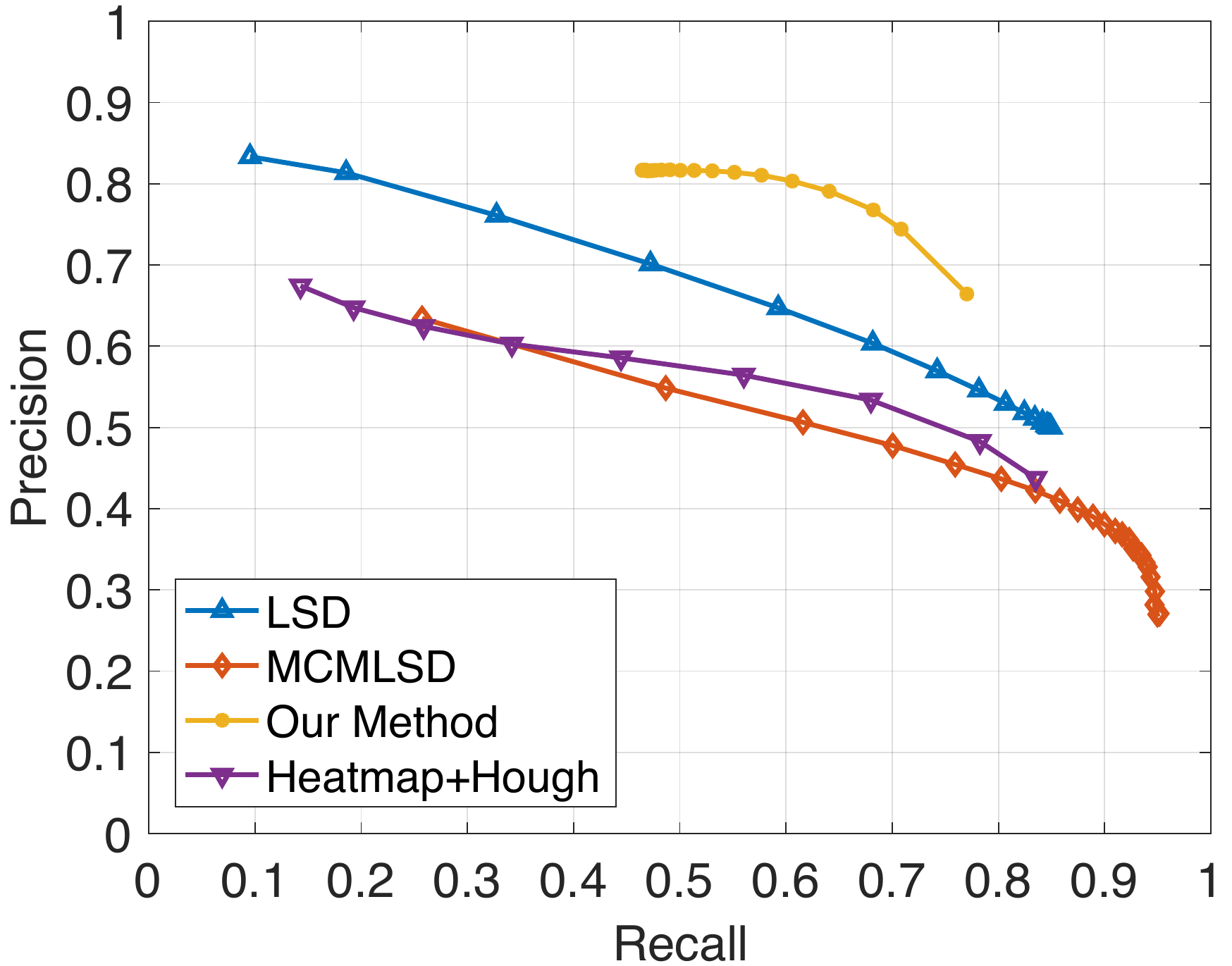} \\
		(a) Our test dataset \\
		\includegraphics[height = 1.8in]{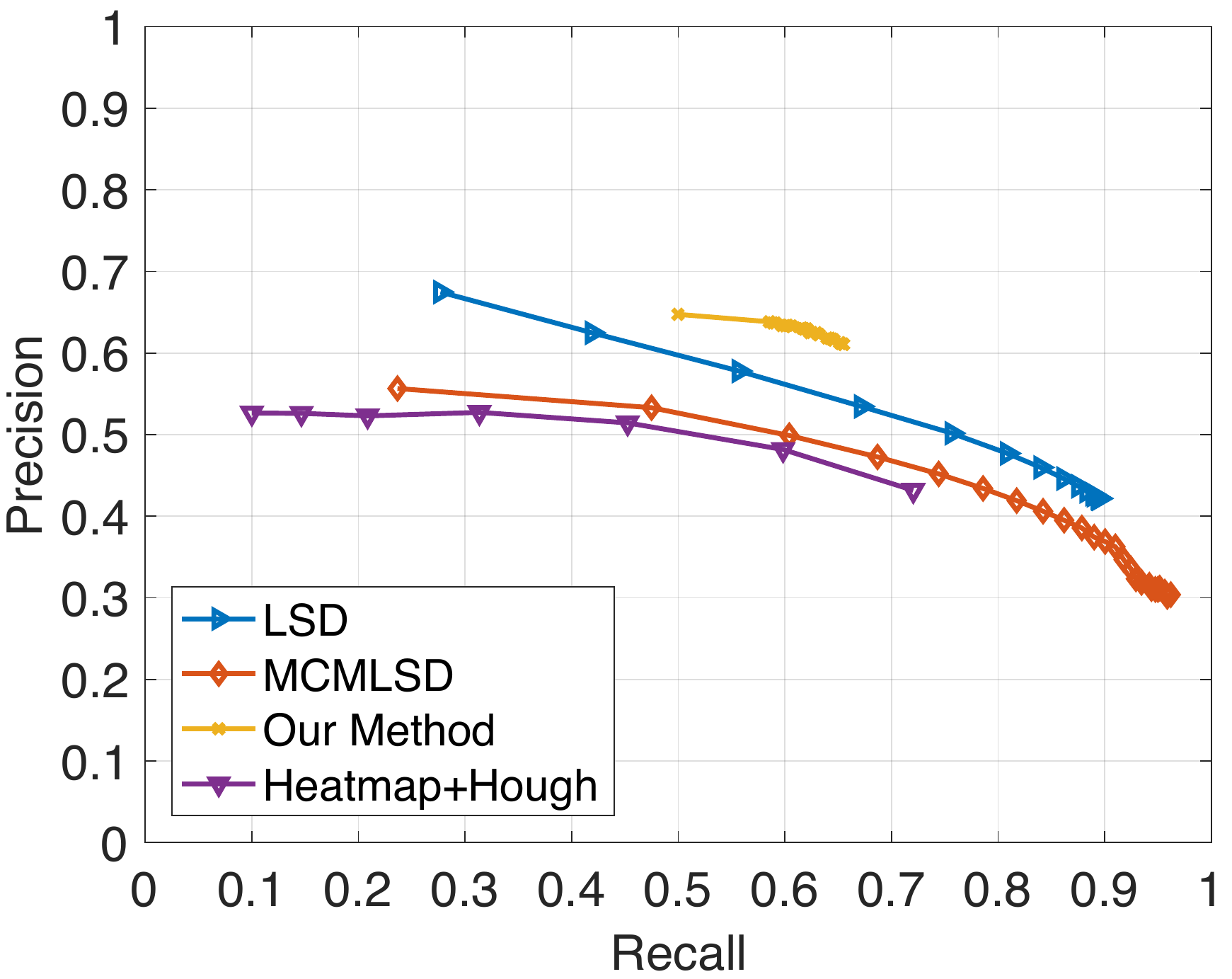} \\
		(b) York Urban dataset
	\end{tabular}
	\caption{Experiment on line segment detection.}
	\label{fig:hough}
\end{figure}

Figure~\ref{fig:hough} shows the precision-recall curves of all methods. We make the following observations on the results: {\em First}, the performance of our ``Heatmap + Hough'' approach is comparable to that of the state-of-the-art line segment detection method MCMLSD, verifying the effectiveness of the our line detection network. {\em Second}, by combining the predicted junctions with the line heat map, our full wireframe construction method performs significantly better than using the line heat map alone. This further illustrates the importance of junction detection in parsing the wireframe: By detecting the ``endpoints'' of the line segments, we effectively overcome the difficulties faced by traditional line segment detection methods, including the false detection problem and the inaccurate endpoint problem.

\subsection*{B.3. Additional Results on Junction Detection}
\begin{figure*}[t]
\centering

\includegraphics[height=0.9in]{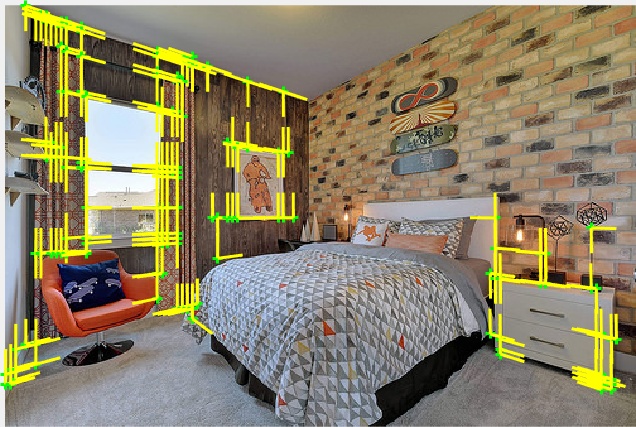}
\includegraphics[height=0.9in]{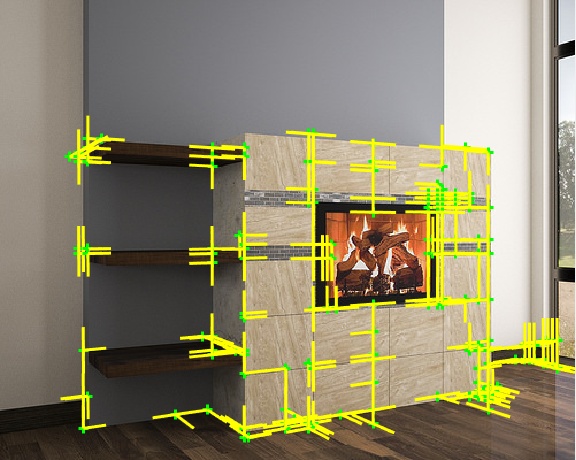}
\includegraphics[height=0.9in]{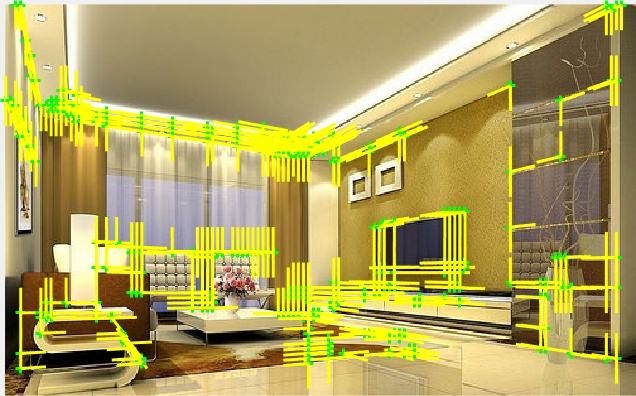}
\includegraphics[height=0.9in]{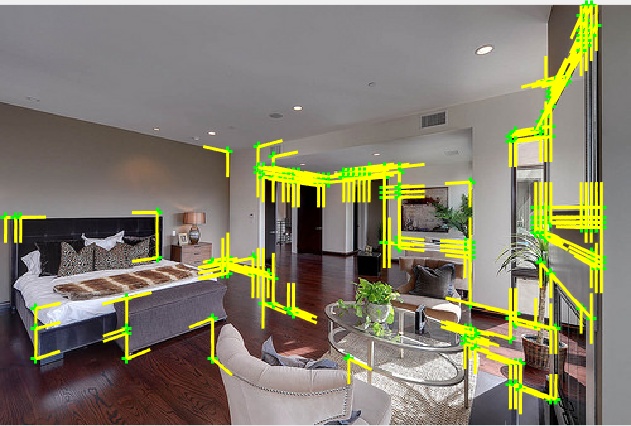}
\includegraphics[height=0.9in]{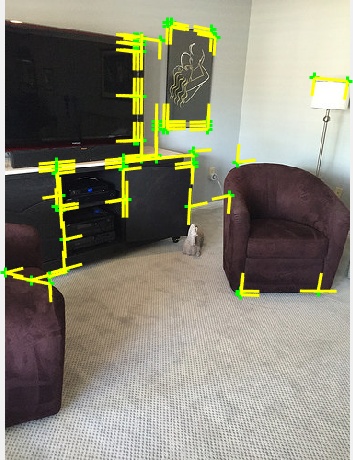}
\\
\includegraphics[height=0.9in]{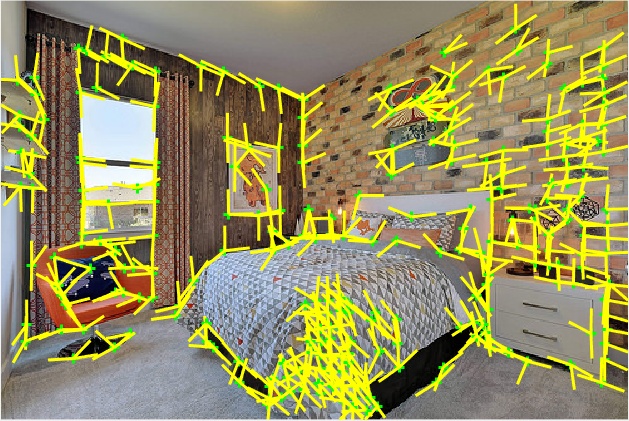}
\includegraphics[height=0.9in]{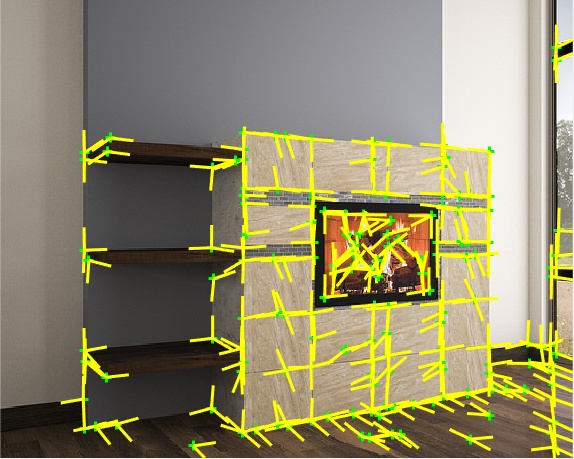}
\includegraphics[height=0.9in]{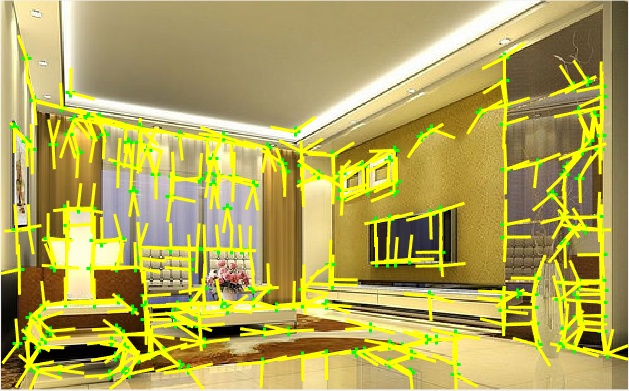}
\includegraphics[height=0.9in]{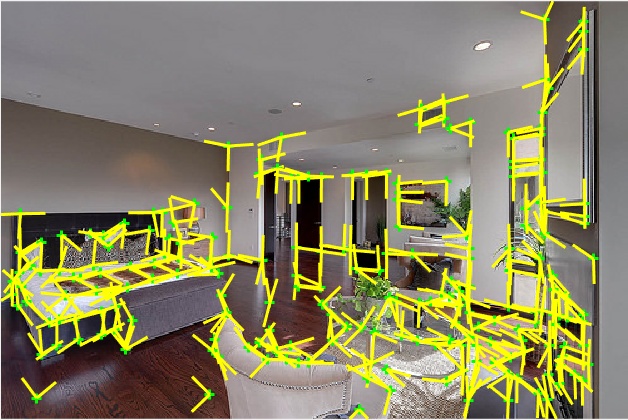}
\includegraphics[height=0.9in]{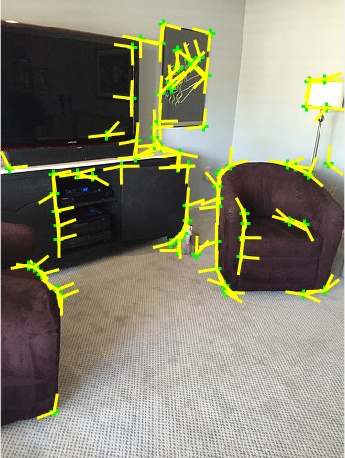}
\\
\includegraphics[height=0.9in]{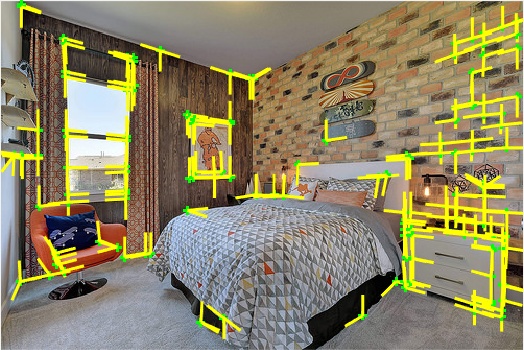}
\includegraphics[height=0.9in]{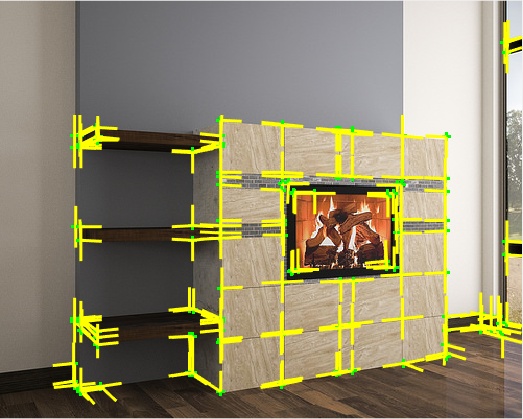}
\includegraphics[height=0.9in]{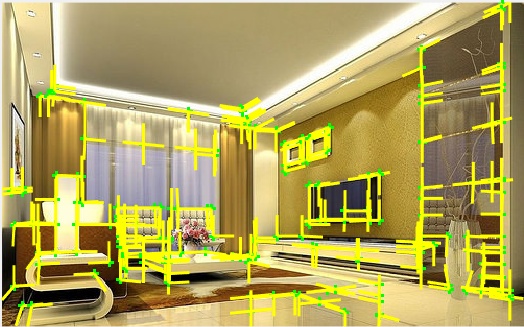}
\includegraphics[height=0.9in]{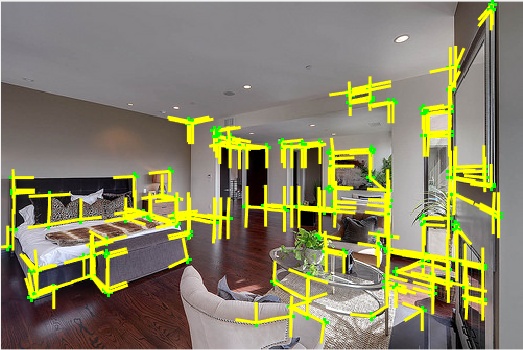}
\includegraphics[height=0.9in]{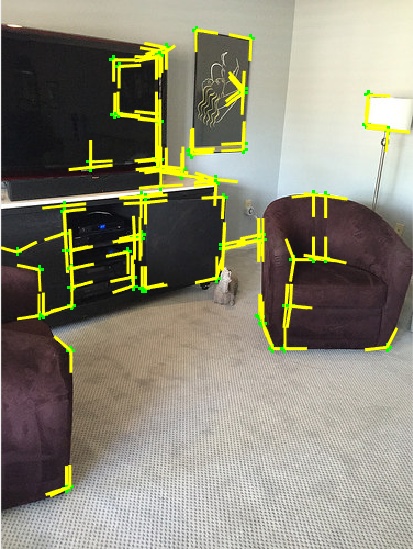}
\\

\includegraphics[height=0.9in]{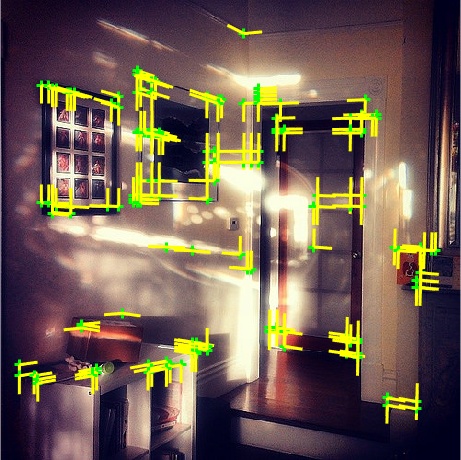}
\includegraphics[height=0.9in]{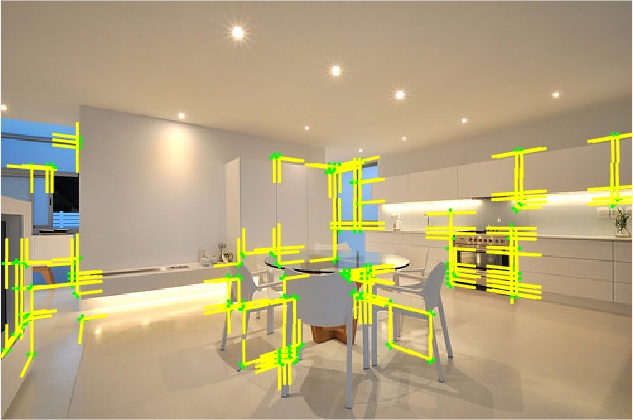}
\includegraphics[height=0.9in]{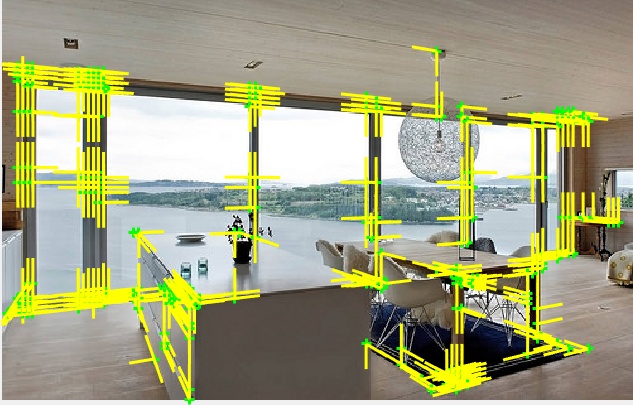}
\includegraphics[height=0.9in]{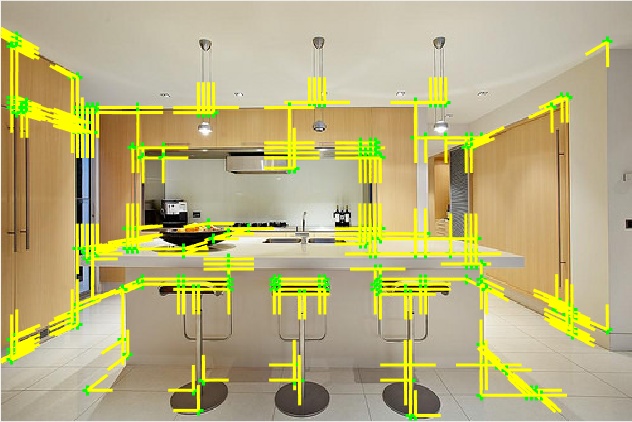}
\includegraphics[height=0.9in]{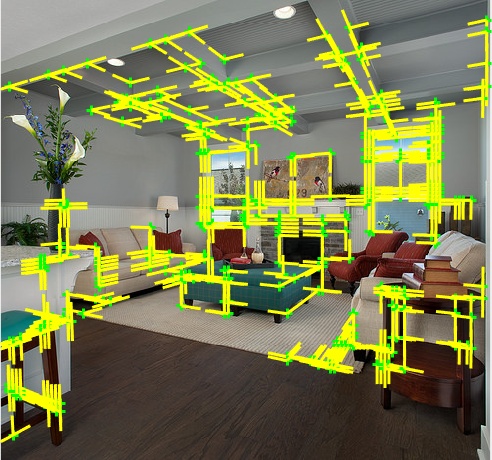}
\\
\includegraphics[height=0.9in]{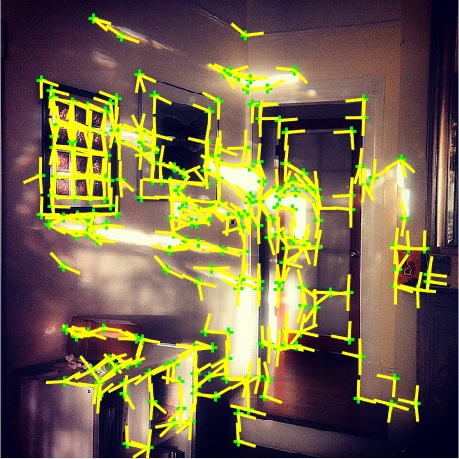}
\includegraphics[height=0.9in]{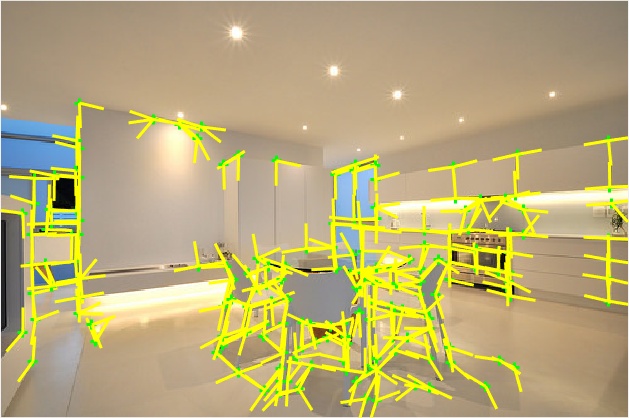}
\includegraphics[height=0.9in]{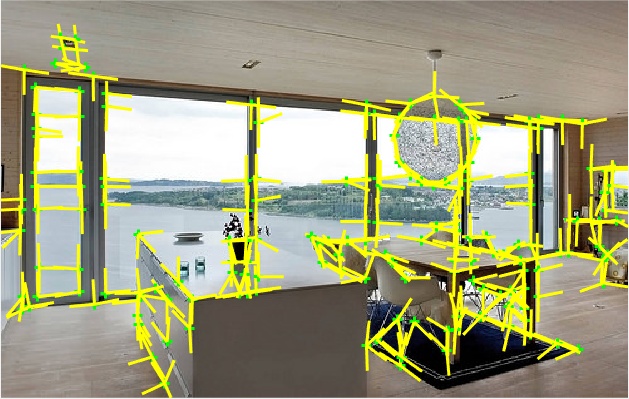}
\includegraphics[height=0.9in]{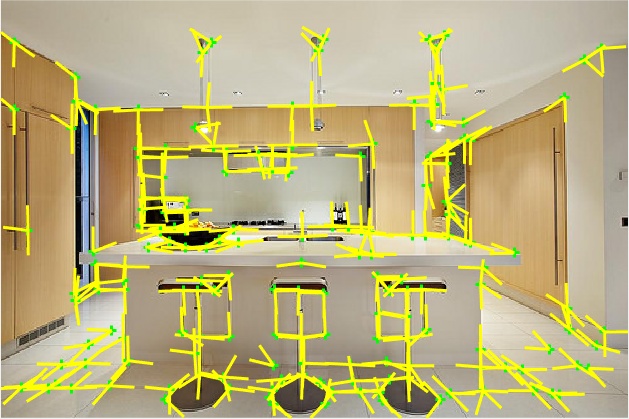}
\includegraphics[height=0.9in]{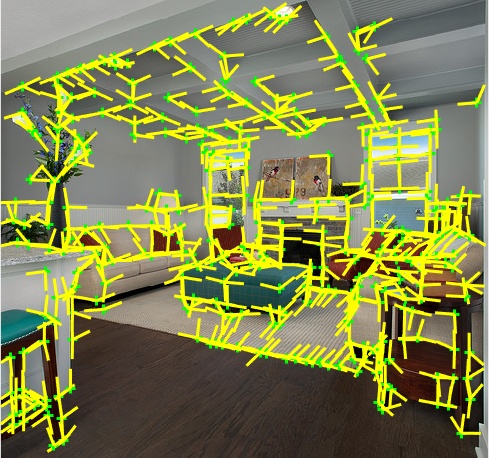}
\\
\includegraphics[height=0.9in]{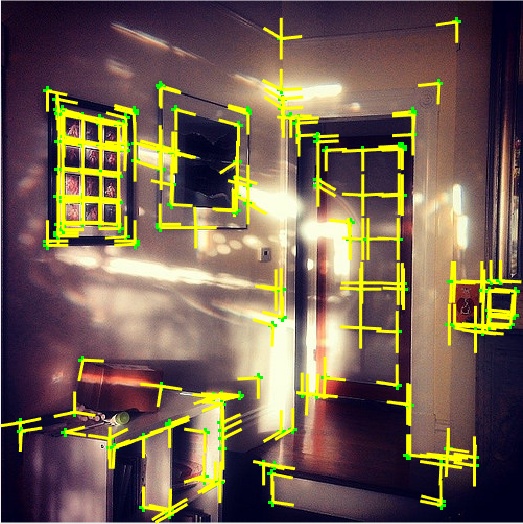}
\includegraphics[height=0.9in]{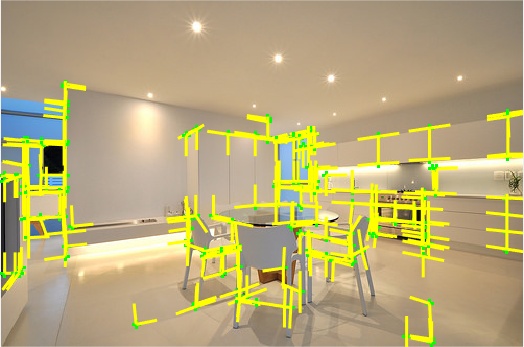}
\includegraphics[height=0.9in]{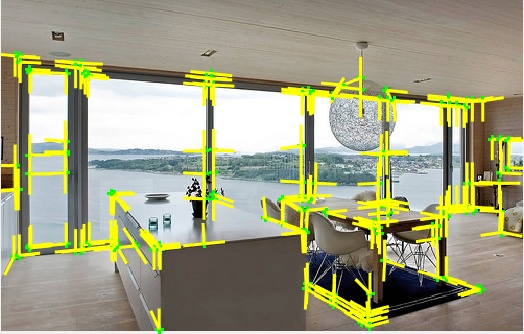}
\includegraphics[height=0.9in]{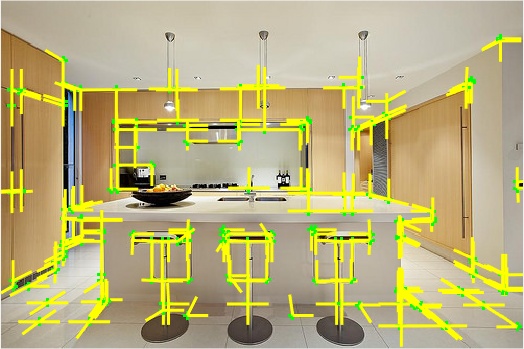}
\includegraphics[height=0.9in]{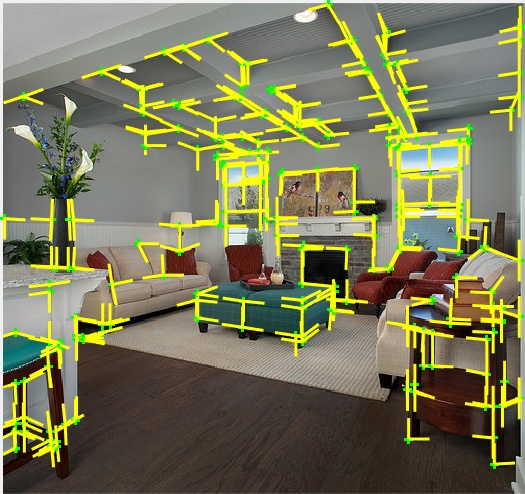}
\\

\includegraphics[height=1.02in]{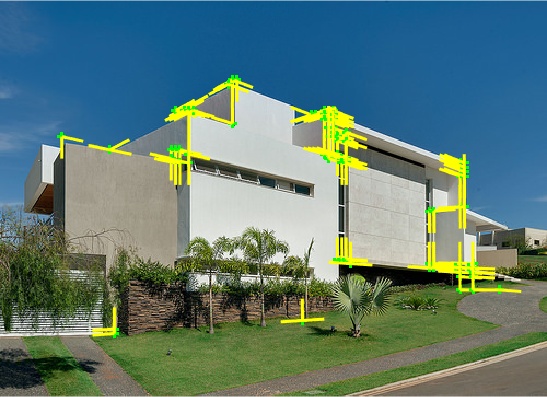}
\includegraphics[height=1.02in]{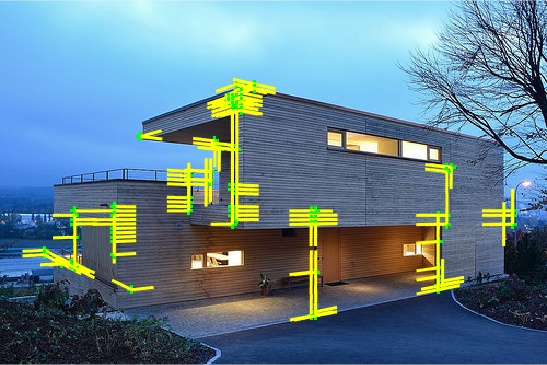}
\includegraphics[height=1.02in]{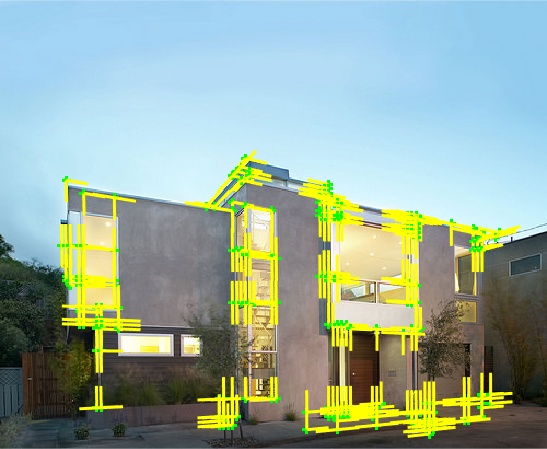}
\includegraphics[height=1.02in]{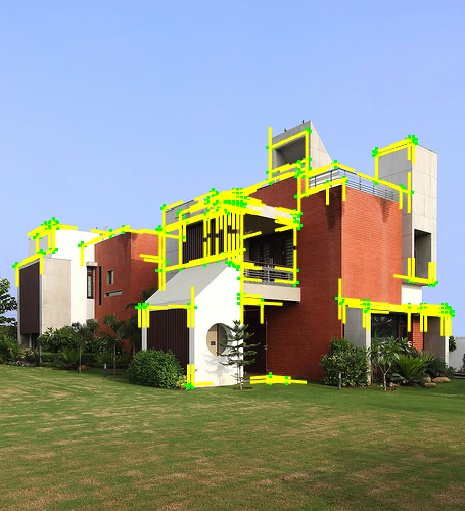}
\includegraphics[height=1.02in]{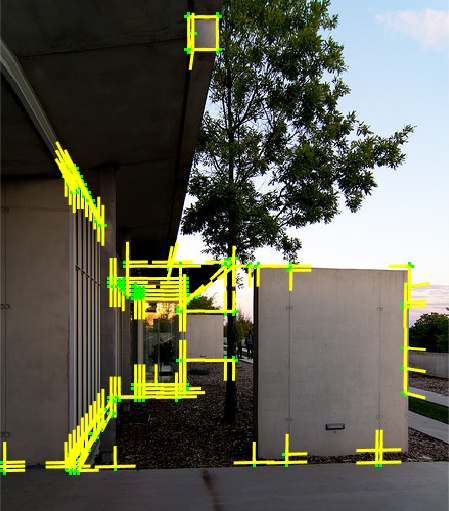}
\\
\includegraphics[height=1.02in]{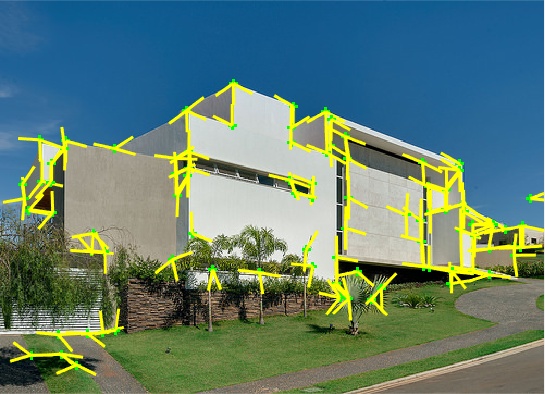}
\includegraphics[height=1.02in]{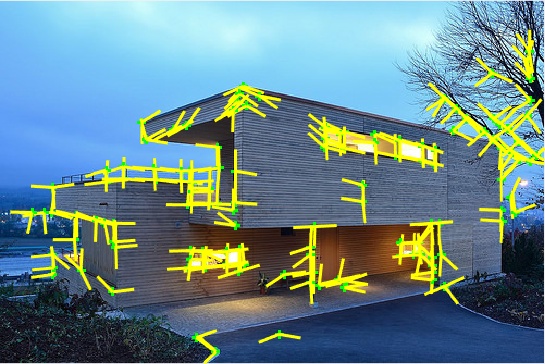}
\includegraphics[height=1.02in]{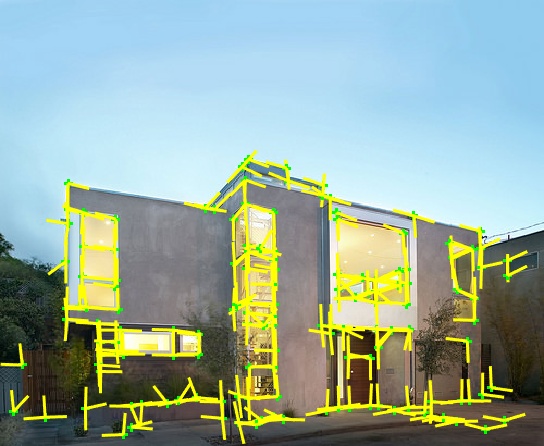}
\includegraphics[height=1.02in]{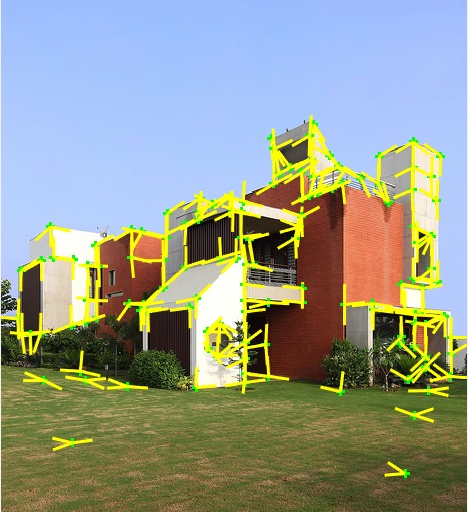}
\includegraphics[height=1.02in]{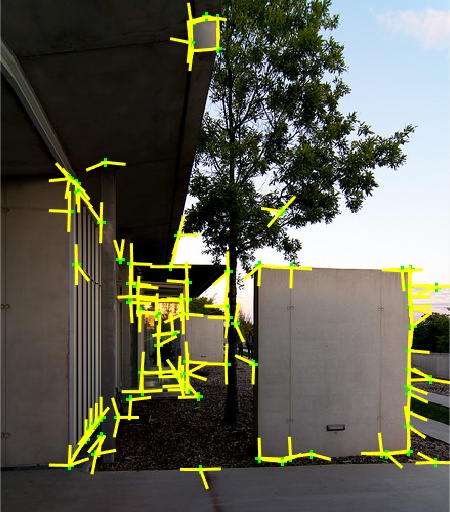}
\\
\includegraphics[height=1.02in]{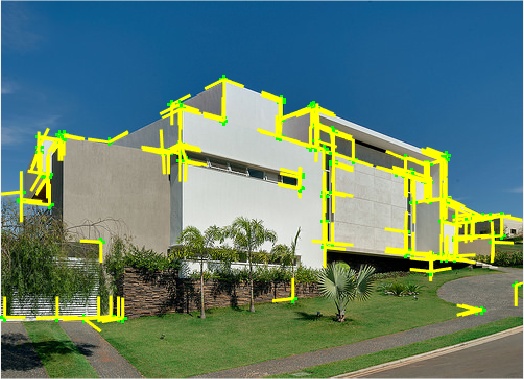}
\includegraphics[height=1.02in]{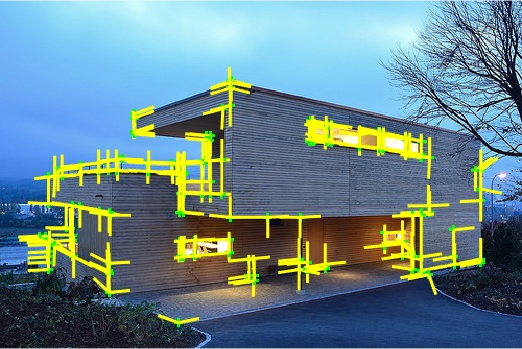}
\includegraphics[height=1.02in]{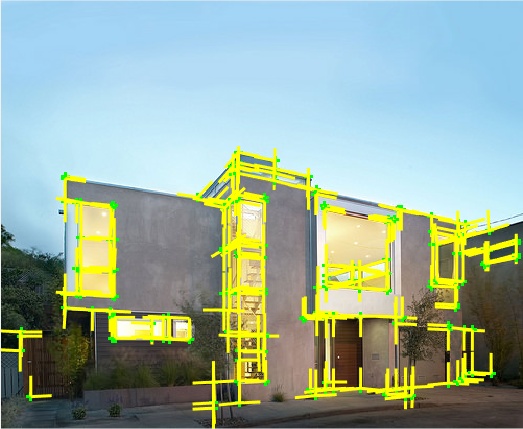}
\includegraphics[height=1.02in]{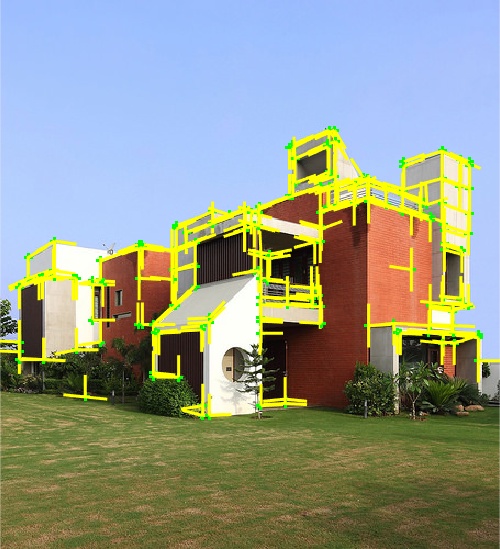}
\includegraphics[height=1.02in]{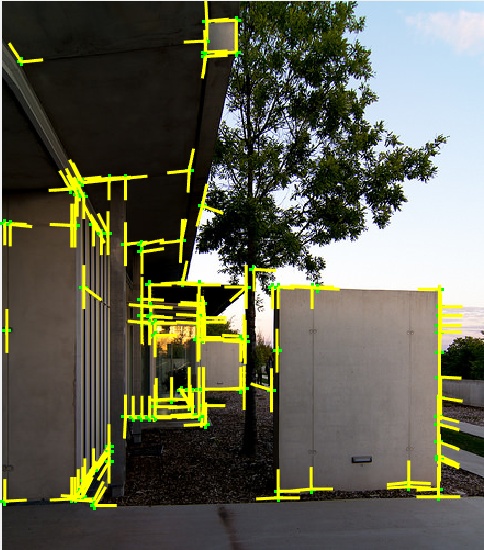}
\\
\caption{Junction detection results. {\bf First row:} MJ ($d_{\max}=20$). {\bf Second row:} ACJ ($\epsilon=1$). {\bf Third row:} Our method ($\tau=0.5$). }
\label{fig:junc-results}
\end{figure*}

In Figure~\ref{fig:junc-results}, we show additional junction detection results obtained by all methods. One can see that our method is able to detect most junctions and their branches in the image, achieving superior performance over existing methods.

From Figure~\ref{fig:junc-results} we can also observe some limitations of our method. Specifically, there are occasionally repeated detections in our result. This may be caused by junctions located at the boundary of two adjacent grid cells used in our junction detection network. Similarly, the use of grid could also lead to missed detection if two junctions are very close to each other. But we note that such cases are rather uncommon in practice and have very small effect on the overall scene structure estimation.

\begin{figure}[t]
	\centering
\includegraphics[height=0.97in]{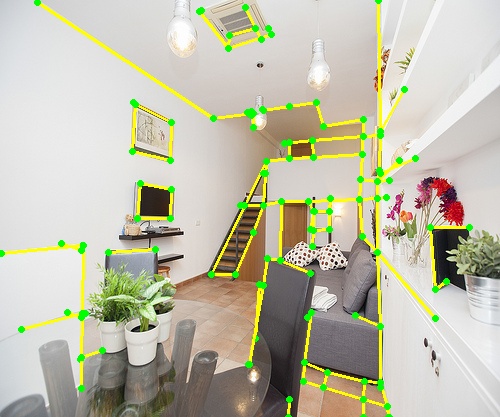}
\includegraphics[height=0.97in]{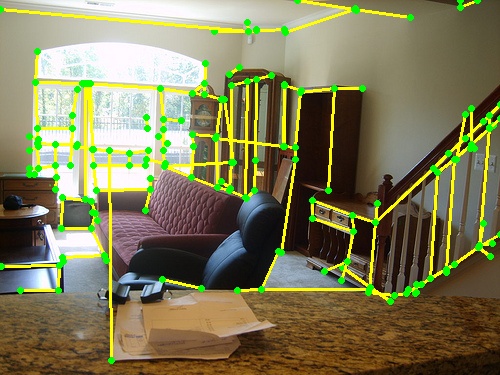}
\includegraphics[height=0.97in]{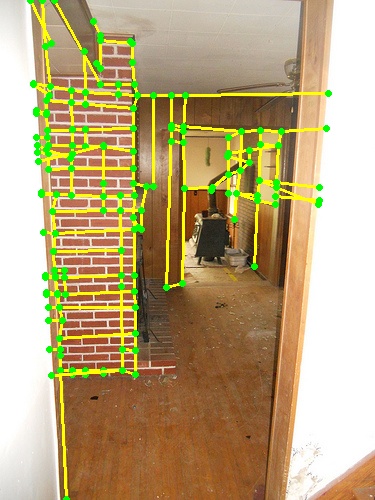}
\\
\includegraphics[height=0.97in]{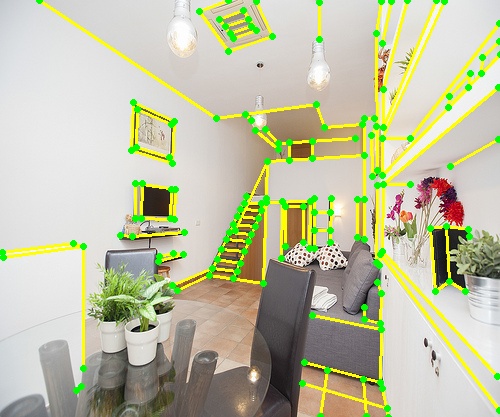}
\includegraphics[height=0.97in]{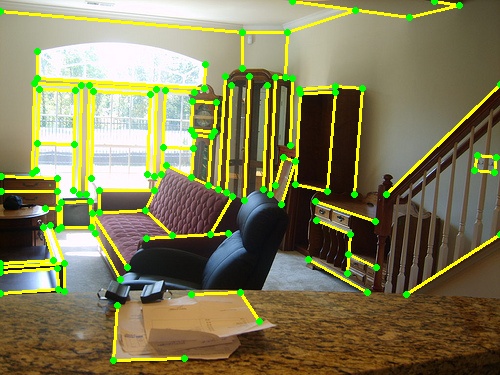}
\includegraphics[height=0.97in]{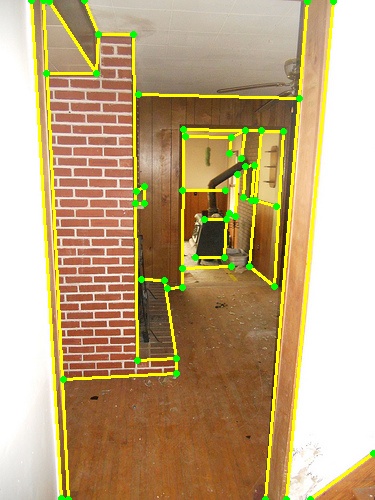}
	\caption{Failure cases on our test dataset. {\bf First row:} Our method. {\bf Second row:} Ground truth.}
	\label{fig:fail}
\end{figure}

\subsection*{B.4. Additional Results on Wireframe Construction}
\begin{figure*}[t]
\centering
\includegraphics[height=1.00in]{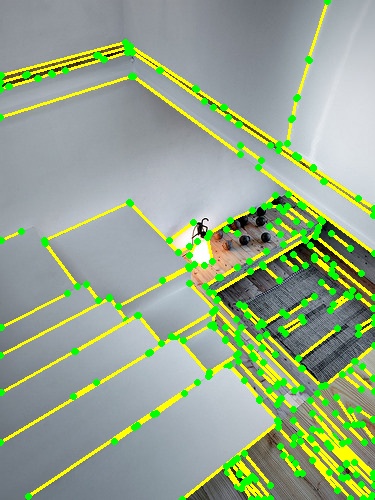}
\includegraphics[height=1.00in]{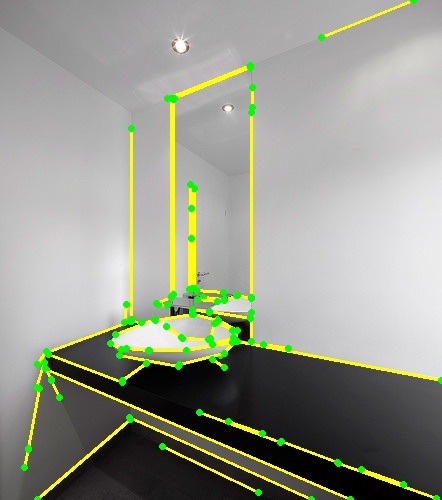}
\includegraphics[height=1.00in]{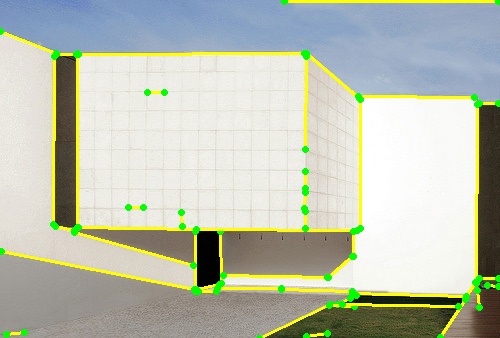}
\includegraphics[height=1.00in]{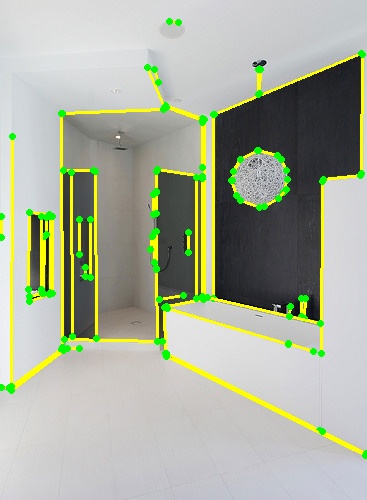}
\includegraphics[height=1.00in]{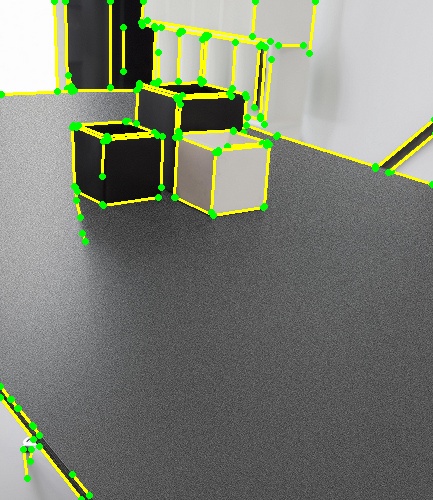}
\includegraphics[height=1.00in]{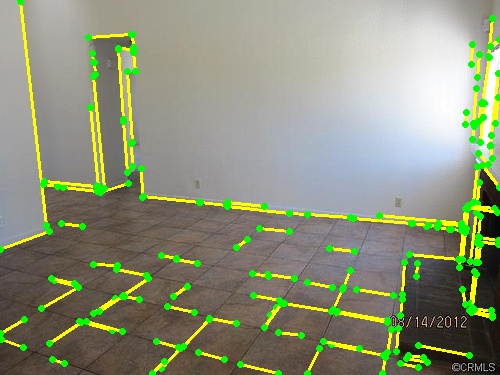}
\\
\includegraphics[height=1.00in]{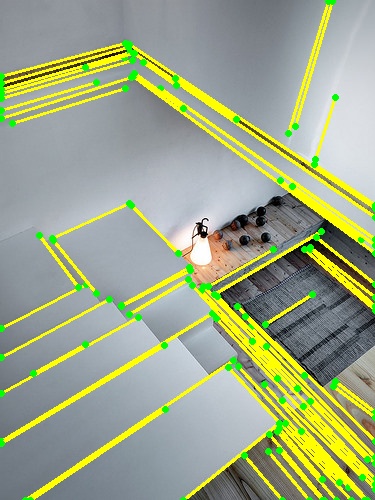}
\includegraphics[height=1.00in]{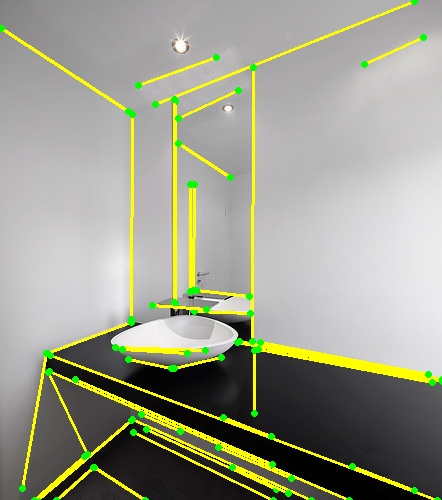}
\includegraphics[height=1.00in]{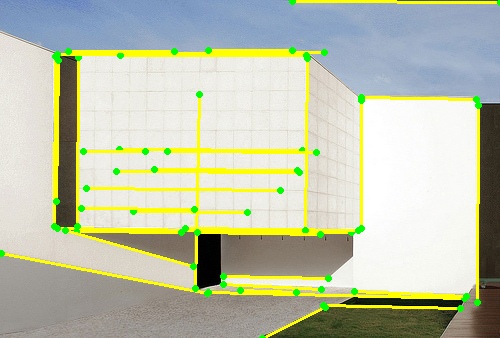}
\includegraphics[height=1.00in]{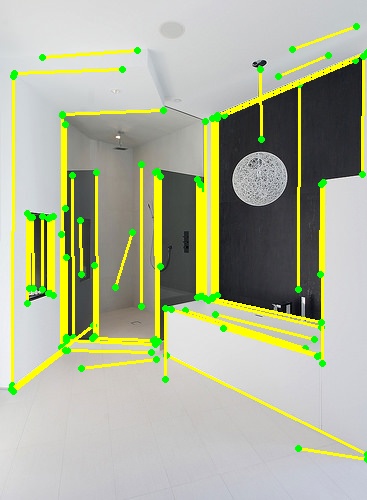}
\includegraphics[height=1.00in]{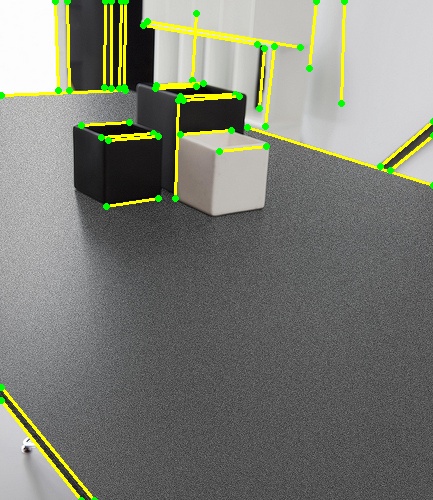}
\includegraphics[height=1.00in]{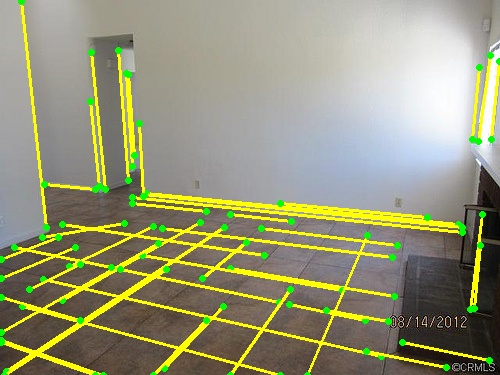}
\\
\includegraphics[height=1.00in]{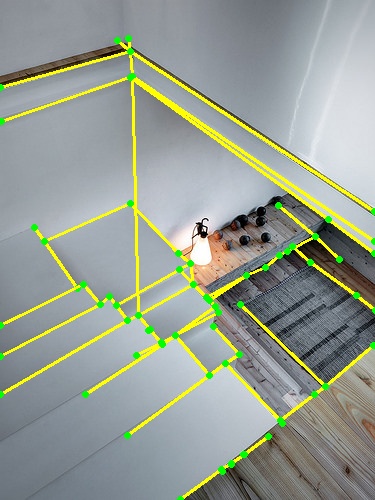}
\includegraphics[height=1.00in]{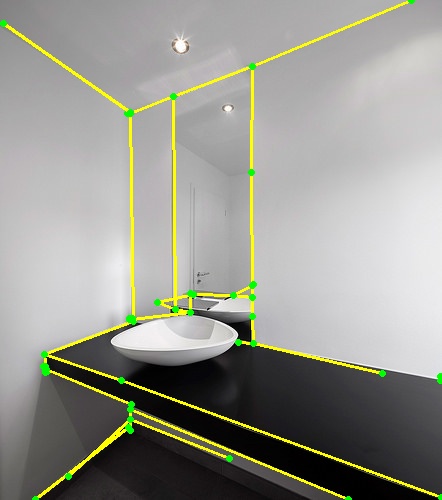}
\includegraphics[height=1.00in]{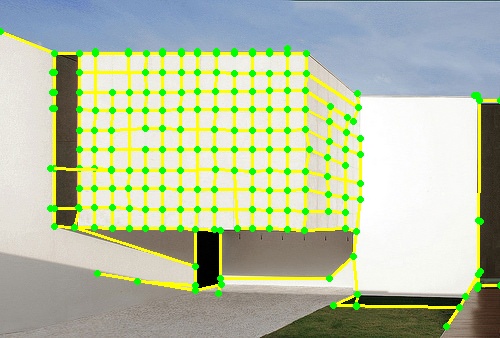}
\includegraphics[height=1.00in]{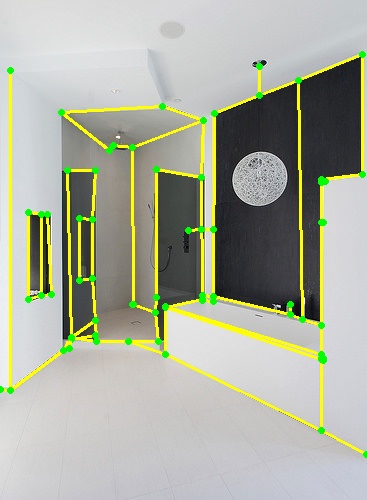}
\includegraphics[height=1.00in]{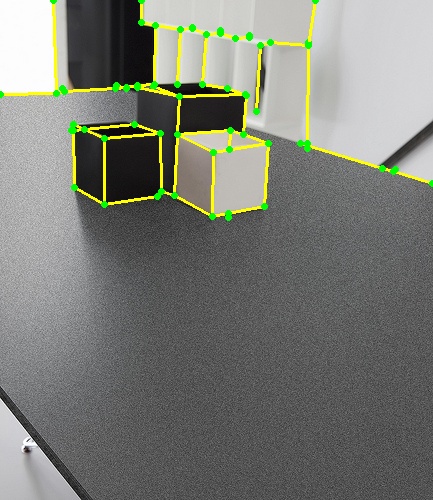}
\includegraphics[height=1.00in]{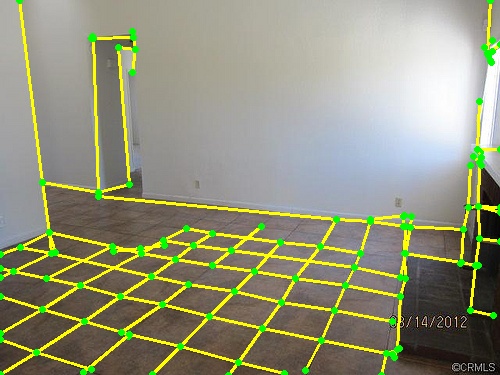}
\\
\includegraphics[height=1.00in]{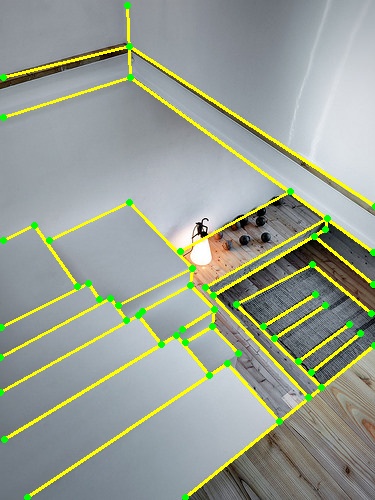}
\includegraphics[height=1.00in]{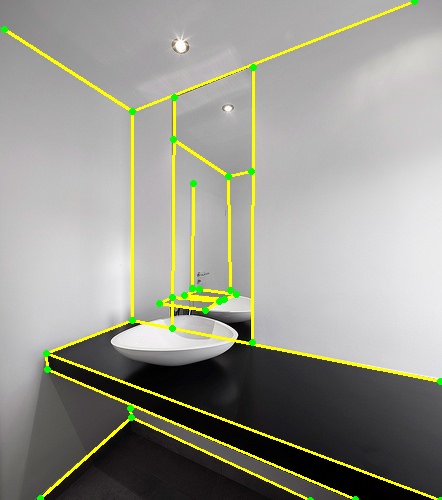}
\includegraphics[height=1.00in]{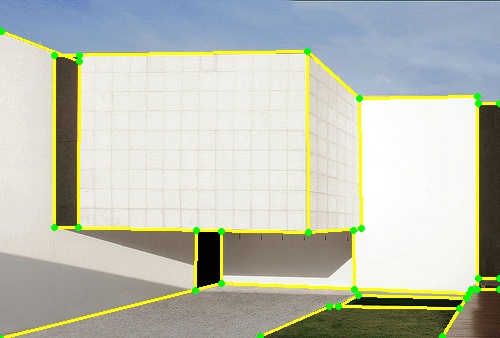}
\includegraphics[height=1.00in]{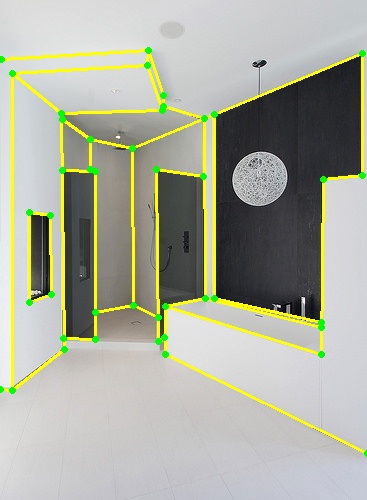}
\includegraphics[height=1.00in]{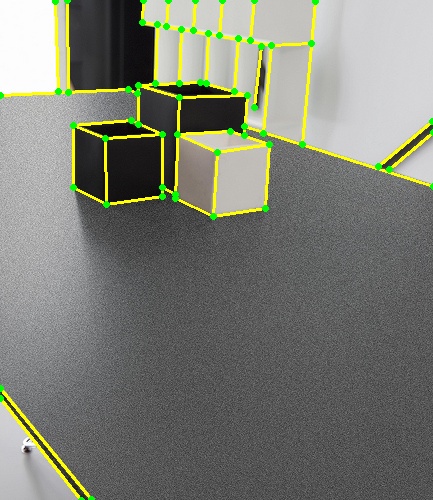}
\includegraphics[height=1.00in]{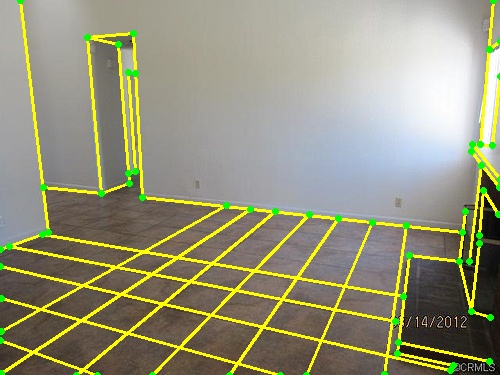}
\\

\includegraphics[height=0.9in]{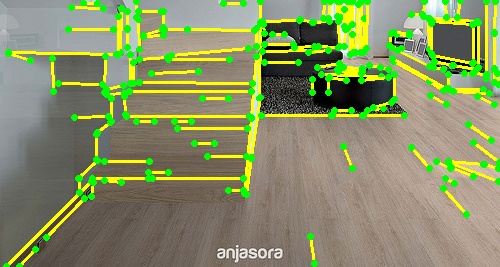}
\includegraphics[height=0.9in]{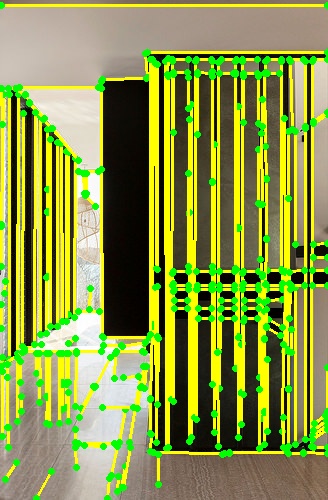}
\includegraphics[height=0.9in]{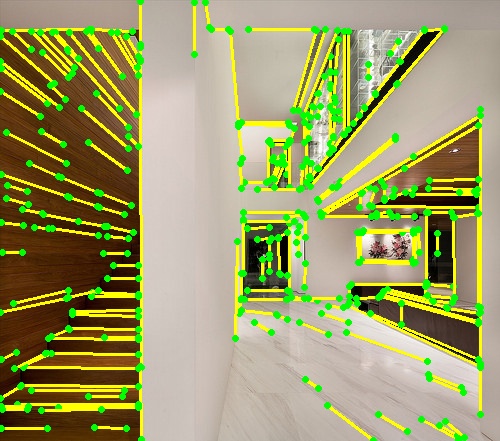}
\includegraphics[height=0.9in]{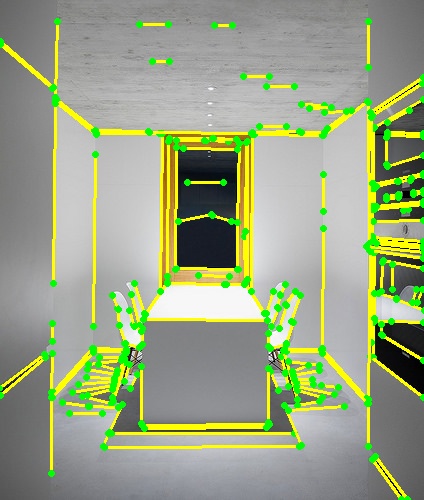}
\includegraphics[height=0.9in]{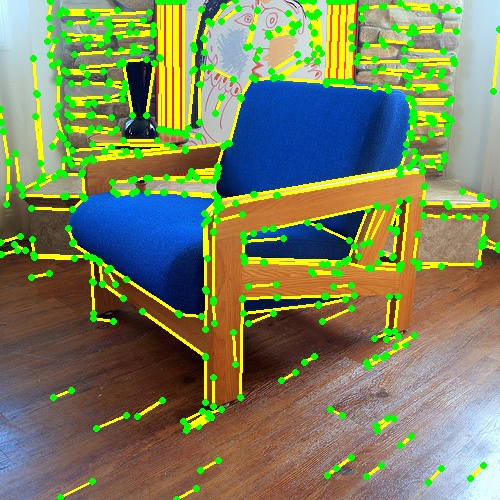}
\includegraphics[height=0.9in]{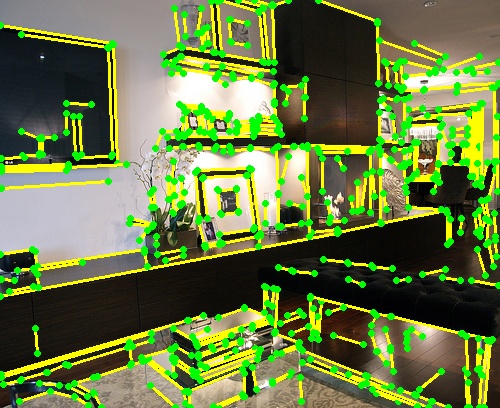}
\\
\includegraphics[height=0.9in]{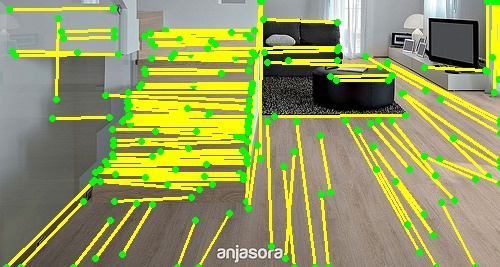}
\includegraphics[height=0.9in]{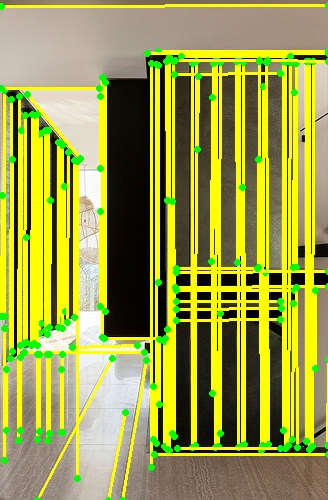}
\includegraphics[height=0.9in]{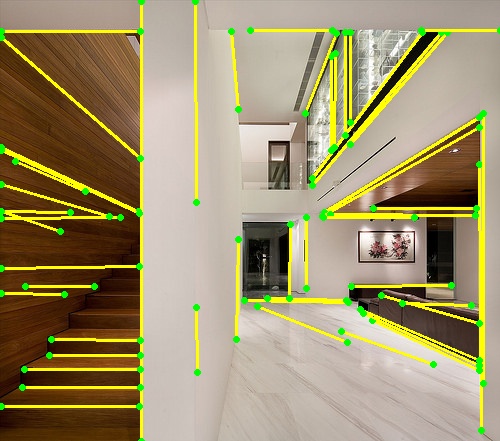}
\includegraphics[height=0.9in]{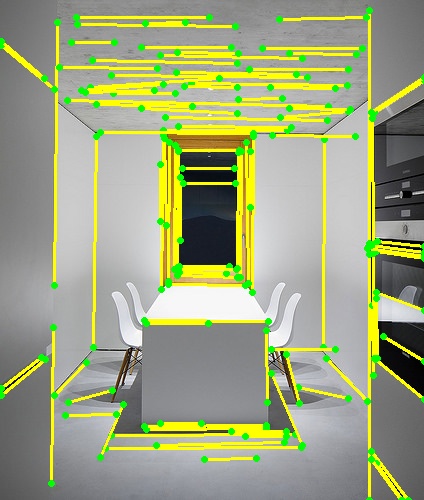}
\includegraphics[height=0.9in]{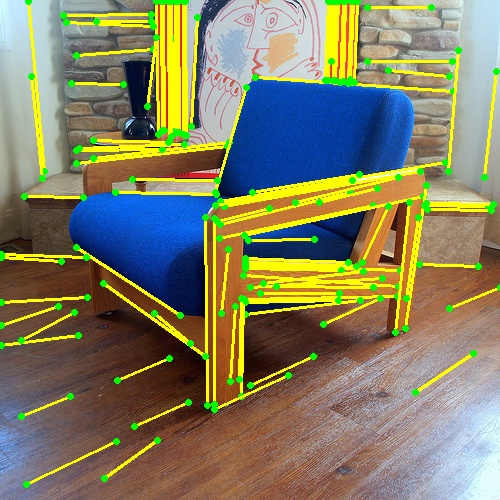}
\includegraphics[height=0.9in]{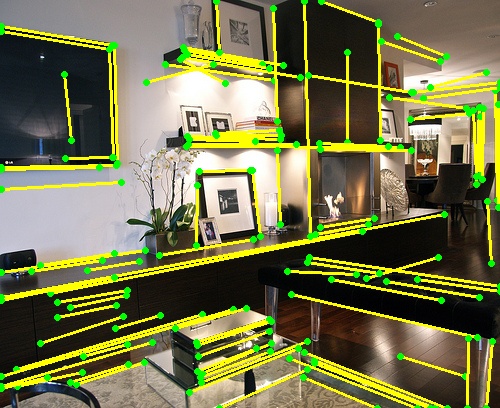}
\\
\includegraphics[height=0.9in]{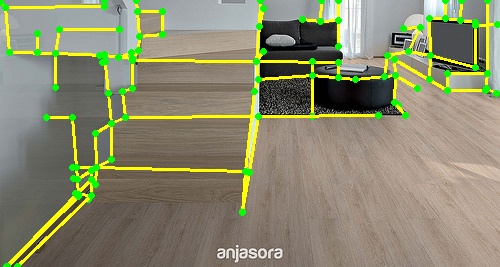}
\includegraphics[height=0.9in]{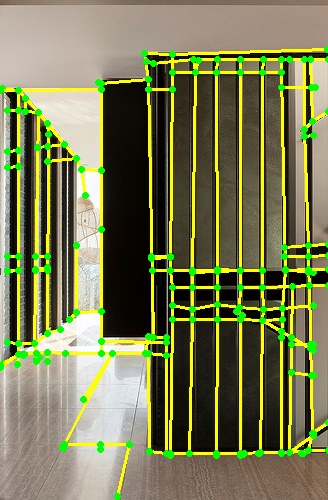}
\includegraphics[height=0.9in]{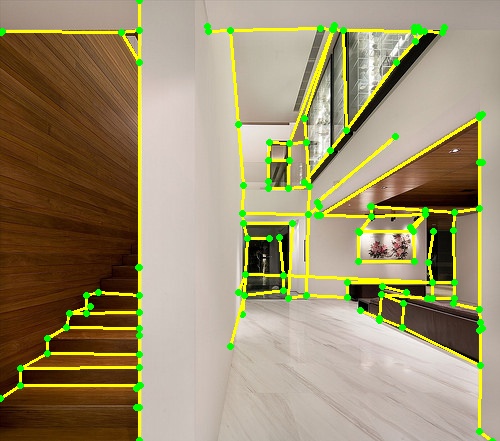}
\includegraphics[height=0.9in]{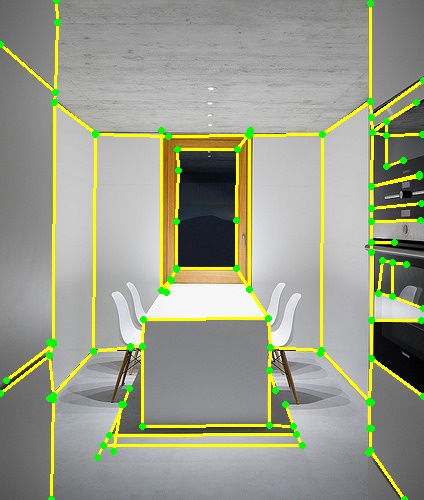}
\includegraphics[height=0.9in]{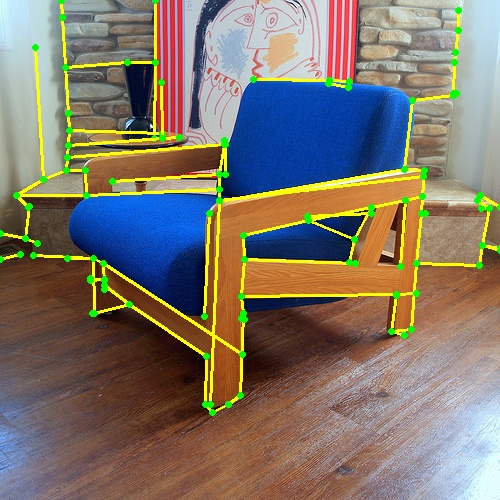}
\includegraphics[height=0.9in]{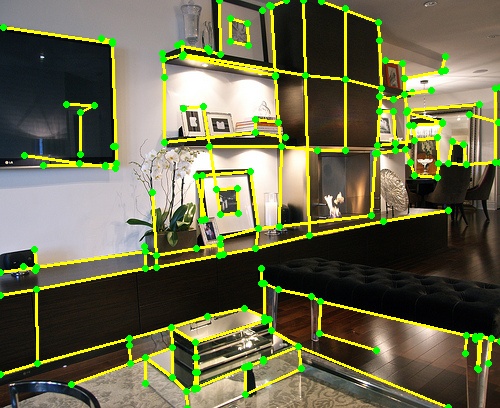}
\\
\includegraphics[height=0.9in]{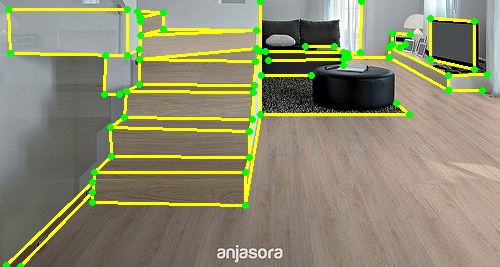}
\includegraphics[height=0.9in]{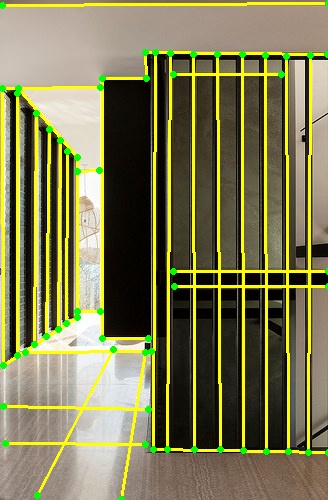}
\includegraphics[height=0.9in]{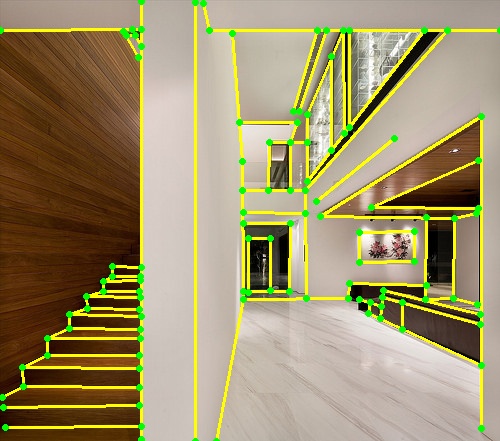}
\includegraphics[height=0.9in]{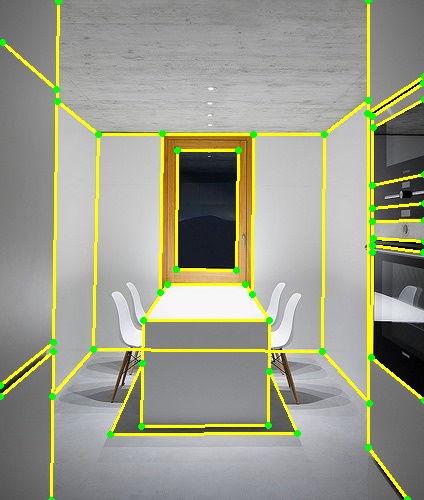}
\includegraphics[height=0.9in]{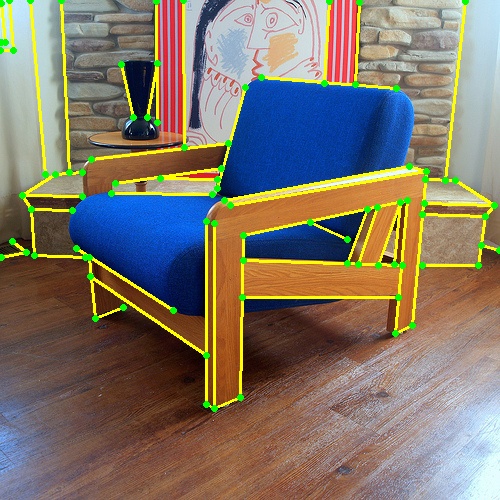}
\includegraphics[height=0.9in]{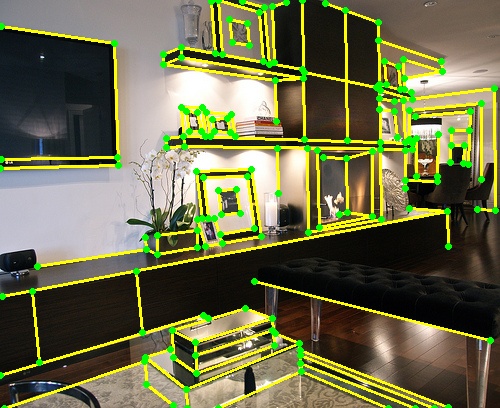}
\\
\caption{Line/wireframe detection results. {\bf First row:} LSD (-$\log$(NFA)  $> 0.01\times1.75^8$). {\bf Second row:} MCMLSD (confidence top 100). {\bf Third row:} Our method (line heat map $h(p) > 10$). {\bf Fourth row:} Ground truth.}
\label{fig:line-results-suppl}
\end{figure*}

In Figure~\ref{fig:line-results-suppl}, we show additional wireframe detection results obtained by all methods. Our method outperforms other two in most areas and produces much cleaner results as we focus on long line segments and exploit their relations (junctions). Therefore, the resulted wireframes are potentially more suitable for 3D reconstruction tasks. 

In Figure~\ref{fig:fail}, we further show some failure cases of our method. One challenging case corresponds to structures with relatively small scale and weak image gradients (e.g., the stairs in the first image). Also, our method sometimes has difficulty in image region of repetitive patterns (e.g., the handrails in the second image and the brick wall in the third image), generating fragment, incomplete results. This suggests opportunities for further improvement by explicitly harnessing such geometric structure in our wireframe construction. 

\end{document}